\documentclass[runningheads]{llncs}

\usepackage{eccv}

\usepackage{eccvabbrv}

\usepackage{graphicx}
\usepackage{booktabs}
\usepackage{multirow}

\usepackage[accsupp]{axessibility}  
\usepackage{graphicx}
\usepackage{booktabs}
\usepackage{multirow}
\usepackage{xcolor}
\usepackage{amsmath,amssymb}
\usepackage{subcaption}
\usepackage{algorithm}
\usepackage{algorithmic}
\usepackage{xspace}
\usepackage{enumitem}
\usepackage{wrapfig}
\usepackage{makecell}
\usepackage{bbm}
\usepackage{tabularx}
\usepackage{color,colortbl}
\usepackage{pifont}

\newcolumntype{Y}{>{\centering\arraybackslash}X}

\usepackage{pgfplots}
\usepgfplotslibrary{groupplots}
\pgfplotsset{compat=1.18}

\usepackage[pagebackref,breaklinks,colorlinks,citecolor=eccvblue]{hyperref}
\usepackage{hyperref}
\hypersetup{
colorlinks = true
}

\usepackage{orcidlink}

\newcommand{\method}{TORA\xspace}
\newcommand{\R}{\mathbb{R}}
\newcommand{\loss}{\mathcal{L}}
\newcommand{\feat}{\mathbf{h}}
\newcommand{\teacher}{\mathbf{y}}

\newcommand{\mainsec}[1]{\textcolor{red}{#1}}

\begin{document}

\title{TORA: Topological Representation Alignment \\ for 3D Shape Assembly} 

\titlerunning{TORA: Topological Representation Alignment}

\author{Nahyuk Lee\inst{1,}$^{*}$ \and
Zhiang Chen\inst{2,}$^{*}$ \and
Marc Pollefeys\inst{2} \and
Sunghwan Hong\inst{2, 3,}$^{\dagger}$}

\authorrunning{N.~Lee et al.}

\institute{Independent Researcher \and
ETH Zurich \and ETH AI Center}
\maketitle
\begingroup
\renewcommand\thefootnote{}
\makeatletter
\renewcommand\@makefntext[1]{\noindent#1}
\makeatother
\footnotetext{${}^{*}$ Equal contribution}
\footnotetext{${}^{\dagger}$ Corresponding author}
\endgroup

\begin{abstract}
  Flow-matching methods for 3D shape assembly learn point-wise velocity fields that transport parts toward assembled configurations, yet they receive no explicit guidance about which cross-part interactions should drive the motion. We introduce \textbf{TORA}, a topology-first representation alignment framework that distills relational structure from a frozen pretrained 3D encoder into the flow-matching backbone during training. We first realize this via simple instantiation, token-wise cosine matching, which injects the learned geometric descriptors from the teacher representation. We then extend to employ a Centered Kernel Alignment (CKA) loss to match the  \textit{similarity structure} between student and teacher representations for enhanced topological alignment. Through systematic probing of diverse 3D encoders, we show that geometry- and contact-centric teacher properties, not semantic classification ability, govern alignment effectiveness, and that alignment is most beneficial at later transformer layers where spatial structure naturally emerges. TORA introduces zero inference overhead while yielding two consistent benefits: faster convergence (up to $6.9\times$) and improved accuracy in-distribution, along with greater robustness under domain shift. Experiments on five benchmarks spanning geometric, semantic, and inter-object assembly demonstrate state-of-the-art performance, with particularly pronounced gains in zero-shot transfer to unseen real-world and synthetic datasets.
  Project page: \href{https://nahyuklee.github.io/tora}{\texttt{https://nahyuklee.github.io/tora}}
  
\end{abstract}

\section{Introduction}
\label{sec:intro}
\begin{figure}[t]
    \centering
    \includegraphics[width=0.98\textwidth]{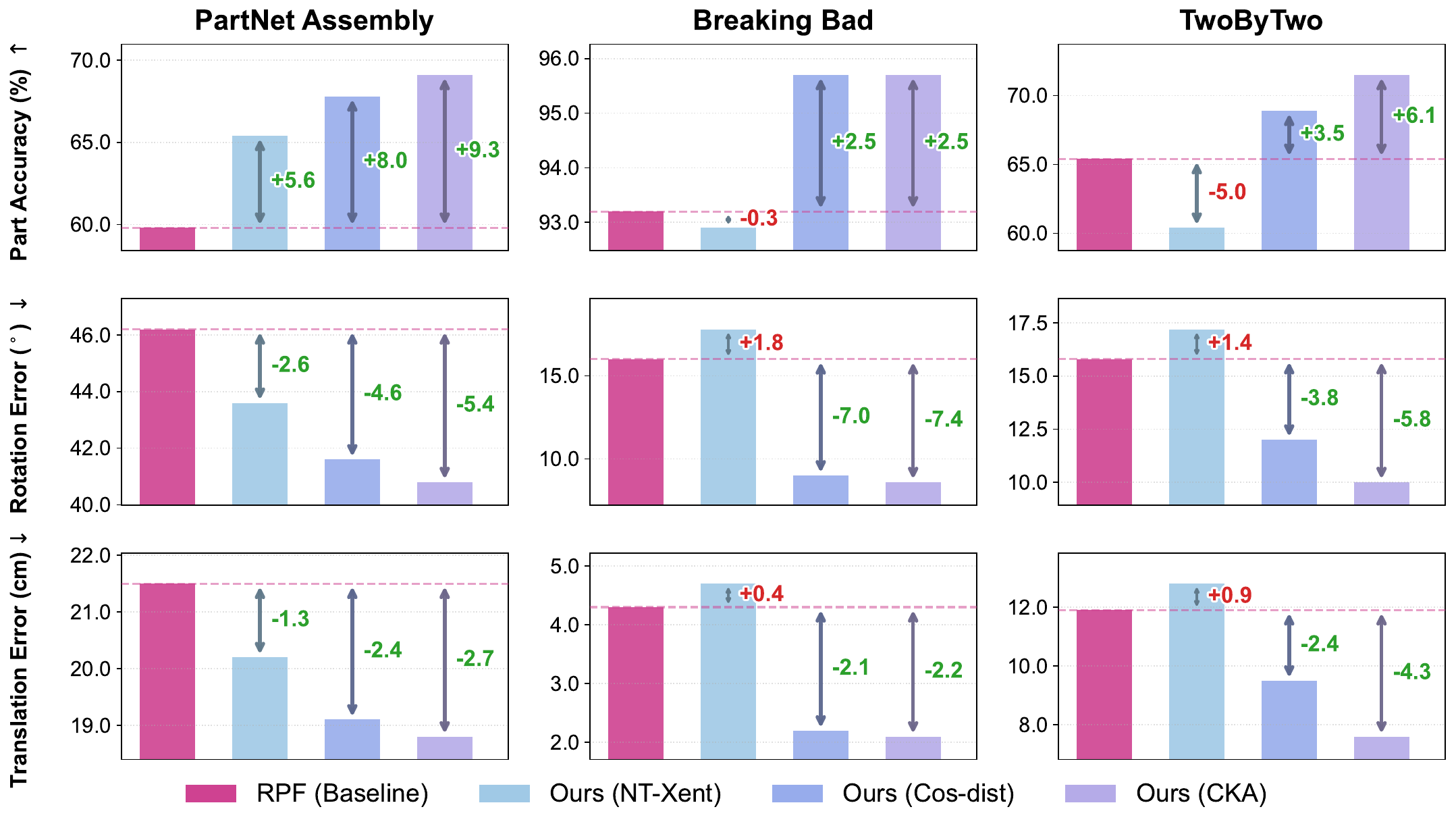}
    \vspace{-1mm}
    \caption{\textbf{Multi-part assembly results across regimes.} We compare the RPF baseline with our alignment variants. We show that casting the training to a teacher-student distillation for injecting pretrained geometric priors consistently improves performance. }\vspace{-15pt}
    \label{fig:quantitative}
\end{figure}

3D object assembly from unposed part point clouds is a fundamental geometric reasoning task with applications in archaeology~\cite{mcbride2003archaeological, son2013axially, yoo2025structure}, computer graphics~\cite{jones2020shapeassembly,chaudhuri2011probabilistic}, and robotics~\cite{harada2016proposal,zakka2020form2fit,qi2025two,chen2026prose,han2026geometric}. The goal is to estimate per-part rigid transformations that reconstruct an object from its constituent parts. Importantly, assembly spans a broad spectrum: semantic assembly, which arranges functionally meaningful intra-object parts (\eg, chair legs and backrests), geometric assembly, which reconstructs physically fractured fragments based on surface complementarity, and inter-object assembly, where components from distinct objects must be mated under cross-object constraints (\eg, peg-in-hole insertion). Across this spectrum, the dominant cues vary from part semantics to fine-grained surface compatibility, but a shared bottleneck is discovering \textit{mating relations}—which regions should contact and co-move—robustly under ambiguity and domain shift.

Motivated by this, recent flow-matching methods~\cite{sun2025rectified,li2025garf} have made significant progress in this direction. In particular, Rectified Point Flow (RPF)~\cite{sun2025rectified} learns a point-wise velocity field that transports noisy point clouds toward assembled configurations, followed by closed-form Procrustes/SVD recovery. While powerful, RPF is trained mainly with an endpoint loss on the final assembled geometry. As a result, the model must infer mating relations implicitly: decisive cues lie on sparse contact regions and are often ambiguous under symmetry, yet the loss does not indicate which cross-part interactions should drive the motion. This lack of explicit intermediate guidance can reduce robustness under distribution shift, motivating the injection of priors that highlight likely interactions.

A practical way to introduce such priors is to cast training in a teacher–student distillation framework, inspired by the 2D generative literature~\cite{yu2024representation} but adapted to 3D point-flow assembly. In our setting, we first show that our simple instantiation, \textit{e.g.,} token-wise cosine matching, is already a strong and effective alignment strategy: by encouraging each point token to match the teacher’s feature content, it injects the teacher’s learned geometric descriptors into the flow backbone and can implicitly transfer relational cues when the teacher representation is well-structured. However, such independent cosine matching treats tokens independently; it does not explicitly constrain the \emph{relational topology} \footnote{We use ``topological'' to refer to relational inter-token similarity structure (\ie, who is similar to whom), not algebraic topology in the sense of persistence or homology.} among points, and may therefore yield suboptimal distillation when assembly hinges on interaction structure across parts.

In this paper, we introduce \textbf{\method}, a teacher–student alignment framework for 3D point-flow assembly that injects pretrained geometric priors into our flow matching model via a frozen 3D teacher encoder. We start from the alignment with standard token-wise objectives, including cosine matching and a contrastive NT-Xent variant~\cite{leng2025repa}, which aligns student tokens to teacher tokens either by directly matching feature content or by additionally enforcing per-point discriminability through negatives. Extending these baselines, we then proceed to introduce a topology-first objective based on Centered Kernel Alignment (CKA)~\cite{cortes2012cka} that directly matches the pairwise \textit{similarity structure} among point tokens, aiming to explicitly preserve the information about who is similar to whom. Finally, we identify two practical choices that matter in this setting—aligning late-layer representations where global geometric structure emerges, and using the right teachers that encode interaction geometry more than category-level semantics.

We demonstrate the effectiveness of \method on several benchmarks spanning semantic, geometric, and inter-object assembly regimes~\cite{sellan2022breaking,qi2025two,xu2025spaformer}, highlighting the importance of transferring relational structure for robust 3D point-flow assembly. As shown in Fig.~\ref{fig:quantitative}, \method achieves state-of-the-art performance across these benchmarks, and further establishes a new state of the art in zero-shot assembly on additional real-world and synthetic datasets~\cite{lamb2023fantastic, li2025garf}. Finally, we provide extensive ablations and analyses to validate our design choices and clarify when each component is most beneficial.

\section{Related Work}

\noindent\textbf{3D shape assembly.}
Shape assembly aims to predict per-part rigid transformations that reconstruct a complete object from its constituent pieces~\cite{sellan2022breaking, huang2020dgl, wu2023leveraging, lu2023jigsaw, lee2024pmtr, lee2025cmnet, wang2024puzzlefusion++, li2025garf, sun2025rectified, xu2025spaformer, li2024category, lu2025survey}.
Early learning-based methods formulate this as a correspondence problem~\cite{cho2021cats,cho2022cats++,an2025cross,an2025c3g,hong2021deep,hong2022cost,hong2022neural,hong2024unifying2,hong2024pf3plat,hong2024unifying,han2025d,han2025emergent}: given point clouds of individual parts, they first establish point-wise correspondences~\cite{lee2024pmtr,lee2025cmnet} across fragments, then extract poses via SVD or weighted averaging~\cite{lu2023jigsaw, lee2024pmtr, lee2025cmnet}.
While effective for pairwise assembly, correspondence-based approaches face combinatorial challenges when scaling to multi-part scenarios, where establishing reliable correspondences across many irregularly shaped fragments becomes increasingly difficult.

Recent work has shifted toward generative formulations that sidestep explicit correspondence estimation~\cite{wang2024puzzlefusion++, li2025garf, sun2025rectified}.
Flow-matching-based methods~\cite{li2025garf, sun2025rectified} learn continuous velocity fields or SE(3) trajectories that transport parts from arbitrary initial poses to their assembled configurations, naturally handling multi-part assembly and part symmetries within a unified probabilistic framework.
Rectified Point Flow (RPF)~\cite{sun2025rectified} achieves the current state-of-the-art by learning a point-wise flow conditioned on features from an overlap-aware encoder, demonstrating strong results across both pairwise registration and multi-part assembly benchmarks.
However, RPF's encoder is pretrained solely on a binary overlap prediction, a relatively narrow geometric signal, and the flow model itself receives no explicit guidance about the broader spatial structure of the parts it assembles.
We show that this geometric understanding can be substantially enriched by distilling representations from pretrained 3D point cloud encoders that have learned richer spatial features from large-scale shape data. 

\medbreak
\noindent\textbf{Distillation objectives for generative models.}
REPA~\cite{yu2024representation,kim2025seg4diff,lee20253d} showed that aligning diffusion transformer~\cite{peebles2023dit} features to a frozen pretrained visual encoder can substantially accelerate training~\cite{yoon2025visual} and improve 2D image generation quality. Subsequent works have expanded this paradigm in several directions: REPA-E~\cite{leng2025repa} enables end-to-end VAE tuning, HASTE~\cite{wang2025repa} augments alignment with attention-based signals and staged termination, and REG~\cite{wu2025representation} introduces discriminative tokens to strengthen denoising. iREPA~\cite{singh2025irepa} further improves spatial fidelity via convolutional projection and spatial normalization, while Geometry Forcing~\cite{wu2025geometry} adds geometric scale and angular constraints for 3D-consistent video diffusion in world-modeling settings. In contrast, our work studies representation alignment for 3D point-flow models in shape assembly, where the target signal is governed by geometric compatibility and sparse mating relations; this leads to different empirical behavior and motivates alignment objectives and teacher choices tailored to part-level 3D geometry.

\section{Method}
\label{sec:method}
\subsection{Problem Formulation}
Given $K$ unposed part point clouds $\{\mathbf{P}_k\}_{k=1}^{K}$ with $\mathbf{P}_k \in \R^{N_k \times 3}$, and $N_k$ is the number of points in part $k \in \{1, \ldots, K\}$, 3D shape assembly seeks per-part rigid transformations $\{\mathbf{T}_k\}_{k=1}^{K}$ with $\mathbf{T}_k \in \text{SE}(3)$ such that the transformed parts $\{\mathbf{T}_k \mathbf{P}_k\}$ reconstruct the original object.
One part is conventionally fixed as an anchor ($\mathbf{T}_1=\mathbf{I}$)\footnote{We discuss the practicality and fairness of this choice and provide additional experiments in the supplementary material.}, and the remaining poses are predicted relative to it.

\subsection{Preliminary: Flow-matching based 3D shape assembly}

\noindent\textbf{Rectified point flow~\cite{sun2025rectified}.}
RPF reformulates this pose estimation problem as conditional generation in 3D Euclidean space.
Rather than directly regressing $\{\mathbf{T}_k\}$, RPF learns a continuous flow that transports points from noise to their assembled positions, from which poses are recovered.
Concretely, let $\mathbf{x}_k(0) \in \R^{N_k \times 3}$ denote the assembled point cloud for part $k$ (sampled from the ground-truth object) and $\mathbf{x}_k(1) \sim \mathcal{N}(\mathbf{0}, \mathbf{I})$.
RPF defines a straight-line interpolation:
\begin{equation}
    \mathbf{x}_k(t) = (1 - t)\,\mathbf{x}_k(0) + t\,\mathbf{x}_k(1), \quad t \in [0, 1],
\end{equation}
with constant velocity $d\mathbf{x}_k(t)/dt = \mathbf{x}_k(1) - \mathbf{x}_k(0)$.
A flow model $V$ is trained to predict this velocity conditioned on the unposed parts $\{\mathbf{P}_k\}_{k=1}^{K}$, which are first encoded by a pretrained overlap-aware encoder.
At inference, the learned flow is integrated from $t{=}1$ (noise) to $t{=}0$ to recover assembled positions $\hat{\mathbf{x}}_k(0)$, and per-part poses are extracted via Procrustes alignment:
\begin{equation}
    \label{eq:procrustes}
    \hat{\mathbf{T}}_k = \arg\min_{\mathbf{T} \in \text{SE}(3)} \|\mathbf{T}\,\mathbf{P}_k - \hat{\mathbf{x}}_k(0)\|_F.
\end{equation}
The training objective supervises the flow through the endpoint reconstruction of $\mathbf{x}_k(0)$, without explicitly specifying which cross-part correspondences or contact regions should drive the flow. Consequently, mating cues—typically sparse and sometimes ambiguous under symmetry—must be discovered implicitly from this global endpoint signal. The flow model processes concatenated point tokens from all parts through a sequence of transformer blocks, producing intermediate representations $\feat^{(l)} \in \R^{N \times D}$ at each layer $l$ (where $N = \sum_k N_k$), which we target for alignment.

\begin{figure}[t]
    \centering
    \includegraphics[width=0.98\textwidth]{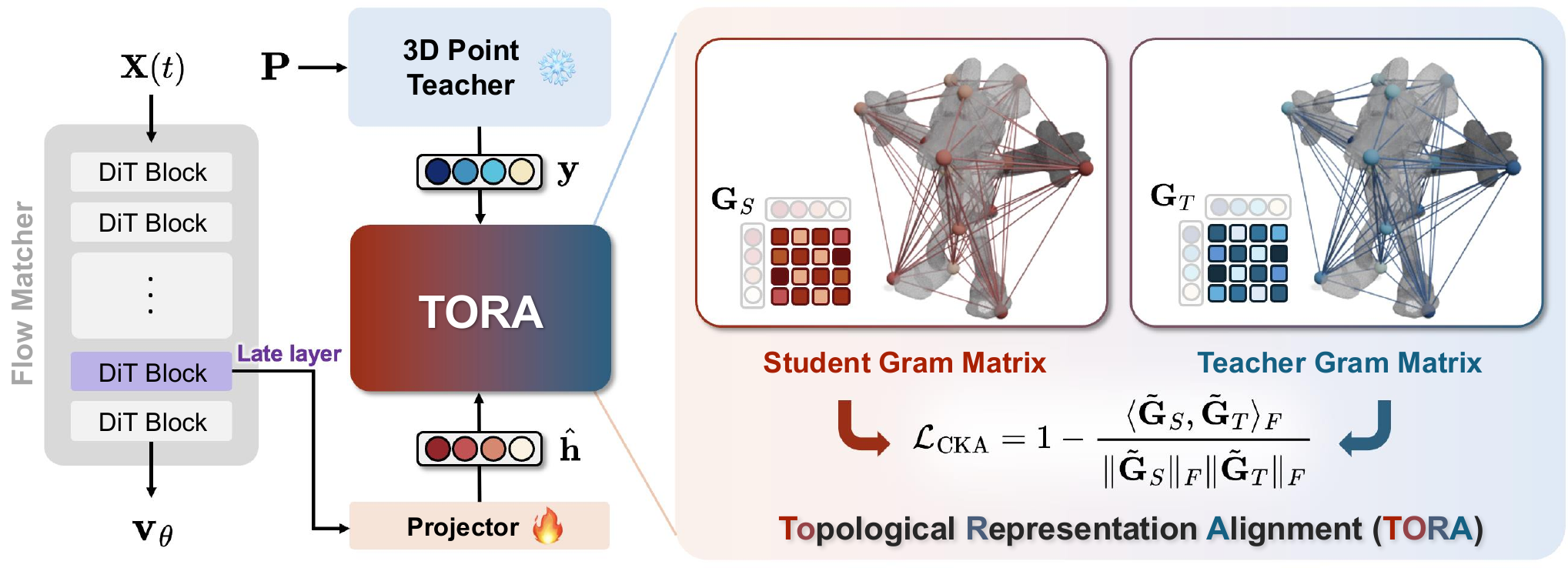}
    \caption{\textbf{Overview of the Topological Representation Alignment (TORA) framework.} TORA distills relational geometric structures from a frozen 3D foundation teacher into a flow-matching student during training. By matching Gram-based similarity matrices via Centered Kernel Alignment (CKA), the student learns the pairwise ``\textit{who-is-similar-to-whom}'' relational topology of parts. As detailed in Sec.~\ref{sec:experiments}, this structural distillation significantly accelerates convergence and enhances robustness under domain shift, while incurring strictly zero overhead during inference.}\vspace{-20pt}
    \label{fig:architecture}
\end{figure}

\subsection{TORA: Topological Representation Alignment}
\label{sec:architecture}

Figure~\ref{fig:architecture} illustrates the overall architecture.
Our methodology is built on RPF architecture~\cite{sun2025rectified}, augmenting its flow matcher with a topological representation alignment branch.

\medbreak
\noindent\textbf{Flow matcher.}
Given $K$ unposed part point clouds $\{\mathbf{P}_k\}_{k=1}^K$, a frozen overlap-aware encoder extracts per-point conditioning features $\mathbf{c} \in \R^{N \times D}$ (see~\cite{sun2025rectified} for details).
A DiT-based~\cite{peebles2023dit} transformer $V_\theta$ takes the noisy point positions $\mathbf{X}(t)$ at timestep $t$ together with $\mathbf{c}$, and predicts a per-point velocity field.
The network consists of $L$ transformer blocks; block $l$ produces intermediate representations $\feat^{(l)} \in \R^{N \times D}$, where $N = \sum_k N_k$ is the total number of points.
Training minimizes the conditional flow matching objective:
\begin{equation}
    \loss_{\text{CFM}}(V_\theta) = \mathbb{E}_{t, \mathbf{X}}\left[\|V_\theta(t, \mathbf{X}(t) \mid \mathbf{X}) - \nabla_t \mathbf{X}(t)\|^2\right],
    \label{eq:cfm}
\end{equation}
where $\mathbf{X}(t) = (1-t)\mathbf{x}_1 + t\,\mathbf{x}_0$ interpolates between target assembled positions $\mathbf{x}_1$ and noise $\mathbf{x}_0 \sim \mathcal{N}(\mathbf{0}, \mathbf{I})$, 
so that $\nabla_t \mathbf{X}(t) = \mathbf{x}_0 - \mathbf{x}_1$.

\begin{figure}[t]
    \centering
    \includegraphics[width=0.98\textwidth]{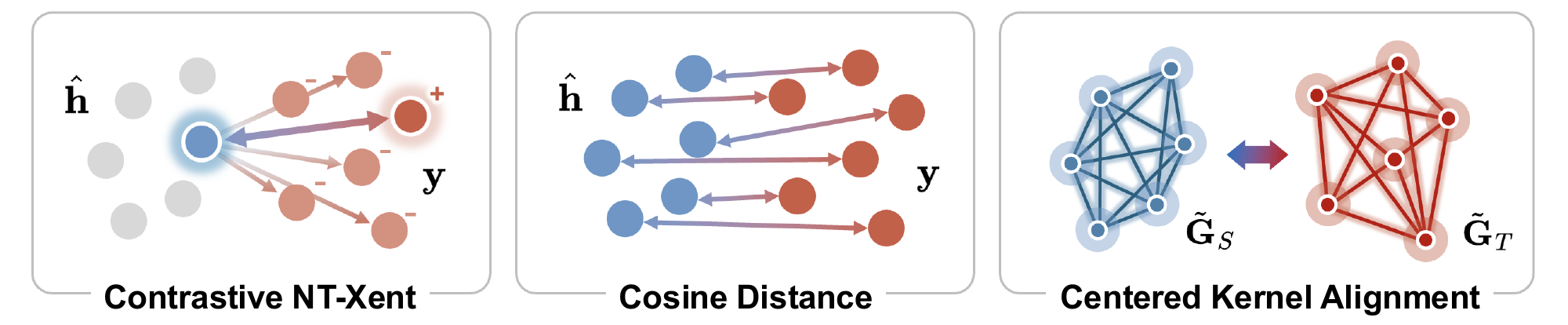}
    \vspace{-1mm}
    \caption{\textbf{Conceptual illustration of alignment objectives.}
  Blue and red dots denote student tokens $\hat{\mathbf{h}}$ and teacher tokens $\mathbf{y}$, respectively.
  NT-Xent enforces per-point discriminability via positive/negative pairing, and cosine distance independently aligns each token pair.
  CKA objective matches the pairwise similarity structures (Gram matrices $\tilde{\mathbf{G}}_S$, $\tilde{\mathbf{G}}_T$), preserving relational topology rather than individual feature vectors.}\vspace{-10pt}
    \label{fig:concept}
\end{figure}

\medbreak
\noindent\textbf{Alignment branch.}
We introduce a frozen \emph{teacher encoder} $f$, selected based on the analysis in Sec.~\ref{sec:analysis}, which produces target representations from the clean (non-noisy) point clouds, $\teacher = f(\mathbf{P}) \in \R^{N \times D_f}$.
A lightweight projector $\phi$ (a 3-layer MLP with SiLU activations~\cite{jocher2021ultralytics}) maps the flow matcher's intermediate features at a selected layer $l^*$ into the teacher feature space:
\begin{equation}
    \hat{\feat} = \phi(\feat^{(l^*)}) \in \R^{N \times D_f}.
\end{equation}
The total training loss augments the conditional flow matching objective with an alignment regularizer,
\begin{equation}
    \loss_{\text{total}} = \loss_{\text{CFM}} + \lambda\,\loss_{\text{align}}(\hat{\feat},\, \teacher),
    \label{eq:total_loss}
\end{equation}
where $\loss_{\text{align}}$ admits multiple instantiations.
We organize them along a single axis, namely the explicitness with which teacher relational structure is transferred, and study three objectives of increasing relational explicitness.
First, token-wise cosine distance ($\loss_{\text{cos-dist}}$) is the simplest baseline, transferring teacher relations implicitly through per-token matching.
Second, a contrastive NT-Xent variant ($\loss_{\text{NT-Xent}}$) additionally enforces per-token discriminability via negatives.
Third, our primary objective, Centered Kernel Alignment ($\loss_{\text{CKA}}$), transfers relations explicitly by matching the pairwise similarity (Gram) structure between student and teacher.
We treat the two token-wise variants as baselines and ablations, and adopt $\loss_{\text{CKA}}$ as the default TORA objective, since it consistently performs best across assembly regimes (Sec.~\ref{sec:experiments}).
We provide a conceptual illustration of the three objectives in Fig.~\ref{fig:concept}.

\medbreak
\noindent\textbf{Token-wise alignment baselines (Cos-Dist, NT-Xent)}. Let $\tilde{\feat}_i = \hat{\feat}_i / \|\hat{\feat}_i\|_2$ and $\tilde{\teacher}_i = \teacher_i / \|\teacher_i\|_2$ denote $\ell_2$-normalized features for point token $i$, and let $\text{sim}(\mathbf{a},\mathbf{b})=\mathbf{a}^\top\mathbf{b}$ denote cosine similarity~\cite{cho2024cat} for normalized vectors. 
We consider (i) a contrastive NT-Xent objective that treats $(\tilde{\feat}_i,\tilde{\teacher}_i)$ as a positive pair and all $(\tilde{\feat}_i,\tilde{\teacher}_j)$ for $j\neq i$ as negatives with temperature $\tau$,
\begin{equation}
\loss_{\text{NT-Xent}}(\hat{\feat},\teacher)
= -\frac{1}{N}\sum_{i=1}^{N}
\log
\frac{\exp\left(\text{sim}(\tilde{\feat}_i,\tilde{\teacher}_i)/\tau\right)}
{\sum_{j=1}^{N}\exp\left(\text{sim}(\tilde{\feat}_i,\tilde{\teacher}_j)/\tau\right)},
\label{eq:ntxent_loss}
\end{equation}
and (ii) token-wise cosine distance,
\begin{equation}
    \loss_{\text{cos-dist}}(\hat{\feat},\teacher)
    = \frac{1}{N}\sum_{i=1}^{N}\left( 1 - \text{sim}(\tilde{\feat}_i,\tilde{\teacher}_i) \right).
    \label{eq:cos_loss}
\end{equation}
These objectives provide effective training-time guidance by transferring teacher structure at the level of individual point tokens. $\loss_{\text{cos-dist}}$ matches per-point feature content, while $\loss_{\text{NT-Xent}}$ additionally enforces per-point discriminability through negatives. Importantly, because assembly is governed by mating relations, token-wise alignment is most helpful when the teacher features implicitly induce a useful similarity structure among points. However, it does not explicitly constrain inter-point relationships and can be less reliable when those relations are subtle. To explicitly transfer such \emph{relational} structure, we adopt a Centered Kernel Alignment (CKA) objective~\cite{cortes2012cka} that matches the pairwise similarity structure among tokens.

\medbreak
\noindent\textbf{Our primary objective: relational alignment via CKA.}
Given student features $\hat{\feat}\in\R^{N\times D_f}$ and teacher features $\teacher\in\R^{N\times D_f}$, we compute $N \times N$ Gram matrices between the features. However, computing exhaustive Gram matrices and using them for loss computations may introduce intractable overheads. We therefore randomly subsample a shared set of $n$ token indices uniformly to keep the $n\!\times\!n$ Gram matrices tractable, and form
\begin{equation}
    \mathbf{G}_S = \hat{\feat}_{\mathcal{I}}\hat{\feat}_{\mathcal{I}}^{\top}, \qquad \mathbf{G}_T = \teacher_{\mathcal{I}}\teacher_{\mathcal{I}}^{\top} \in \R^{n\times n},
\end{equation}
where $\mathcal{I}\!\subset\!\{1,\dots,N\}$ with $|\mathcal{I}|=n\ll N$ and the subscript denotes row selection.
These matrices encode all pairwise inner products between the sampled tokens. We then center them using
\begin{equation}
    \mathbf{H} = \mathbf{I}-\frac{1}{n}\mathbf{1}\mathbf{1}^{\top}, \qquad 
    \tilde{\mathbf{G}}_S = \mathbf{H}\mathbf{G}_S\mathbf{H}, \quad \tilde{\mathbf{G}}_T = \mathbf{H}\mathbf{G}_T\mathbf{H}.
\end{equation}
Finally, we define the CKA loss as the negative normalized alignment between centered Gram matrices:
\begin{equation}
    \loss_{\text{CKA}}(\hat{\feat},\teacher)
    = 1 - \frac{\langle \tilde{\mathbf{G}}_S, \tilde{\mathbf{G}}_T\rangle_F}
    {\|\tilde{\mathbf{G}}_S\|_F\,\|\tilde{\mathbf{G}}_T\|_F},
    \label{eq:cka_loss}
\end{equation}
where $\langle\cdot,\cdot\rangle_F$ denotes the Frobenius inner product.
Unlike token-wise matching, $\loss_{\text{CKA}}$ explicitly preserves ``who is similar to whom'' across all token pairs, and is invariant to isotropic scaling and orthogonal transformations of the feature space~\cite{cortes2012cka}.

\section{Understanding Alignment for 3D Assembly}
\label{sec:analysis}

\begin{figure}[t]
    \centering
    \includegraphics[width=0.98\textwidth]{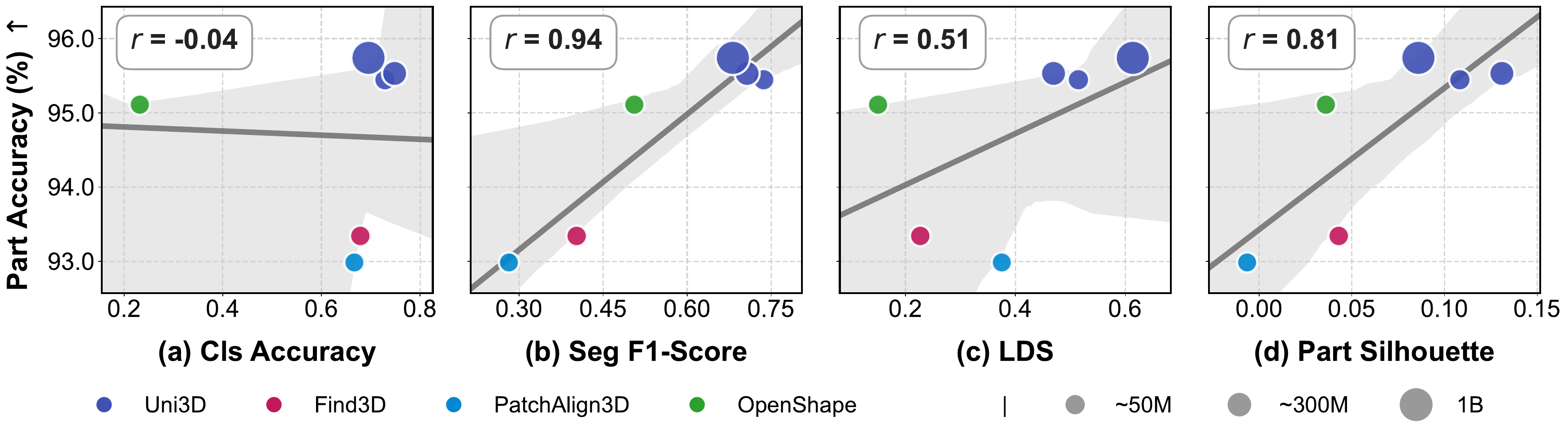}
    \vspace{-2mm}
    
    \caption{\textbf{Correlation Analysis of Teacher Representations.} 
        We analyze the relationship between representation properties and assembly performance. \textbf{(a-b)} Global semantic understanding (object classification) exhibits a much weaker correlation with final shape assembly accuracy in comparison to spatial structure awareness (mating surface segmentation). 
        \textbf{(c-d)} Local spatial structure metrics such as LDS struggle to identify performant teachers, whereas measures highlighting particular geometric properties depict much clearer trends in assembly performance.
        Overall, geometry- and contact-centric teacher properties are more indicative of downstream assembly quality, motivating our structural distillation objective.
    }
    \label{fig:correlation}
\end{figure}

\subsection{What Makes a Good Teacher for 3D Assembly?}
Having defined the alignment objectives above, in this section, we proceed to investigate what makes a good teacher for 3D shape assembly. A key ingredient in our training-time alignment is the choice of a teacher encoder. 
While several pretrained 3D point cloud encoders are available, it is unclear \emph{a priori} which teacher properties are most relevant for 3D part assembly, where success is governed by geometric compatibility and sparse mating interactions. 
To ground this choice, we study a range of pretrained 3D encoders as teachers and ask: which aspects of a teacher representation predict downstream improvements when used for alignment?

To this end, we evaluate six pretrained 3D encoders as teachers~\cite{zhou2023uni3d,hadgi2026patchalign3d,ma2025find,liu2023openshape}. For each teacher model, we compute lightweight linear-probe metrics on frozen features that capture complementary properties: (i) \textbf{classification accuracy} as a proxy for global semantic content, (ii) \textbf{mating-surface segmentation F1} as a proxy for contact-awareness and interaction geometry, (iii) \textbf{ Local-vs-Distant Similarity (LDS)} as a geometry-sensitivity measure, and (iv) \textbf{Part Silhouette} as a proxy for part-level geometric priors.\footnote{We describe probe setups and protocols in the supplementary material.} We then correlate each probe score with downstream assembly performance (Part Accuracy) obtained when the teacher is used for alignment.

Figure~\ref{fig:correlation} shows a consistent trend across the teacher models we evaluate. \textbf{Semantic classification accuracy shows little predictive value} for assembly transfer, exhibiting near-zero correlation with downstream Part Accuracy. In contrast, \textbf{geometry- and interaction-centric probes are more predictive of downstream gains}: mating-surface segmentation F1 shows the strongest association with Part Accuracy, Part Silhouette prediction is also strongly aligned, and LDS exhibits a moderate relationship. Taken together, these results support a simple takeaway: effective teacher supervision for 3D assembly depends more on encoding \emph{interaction geometry}—potential contact regions and shared geometric context across parts—than on category-level semantics.

These observations suggest a practical guideline for teacher selection in 3D shape assembly: teacher quality is better assessed using \textbf{geometry/contact probes} than global recognition metrics. Guided by this, we adopt {Uni3D}~\cite{zhou2023uni3d} as our default teacher, as it consistently achieves strong geometry/contact probe performance and yields the best downstream assembly accuracy across alignment objectives. We use Uni3D throughout the paper unless stated otherwise, and further justify this choice in the following section.

\subsection{Teacher Choice for Distillation}

\begin{wrapfigure}{r}{0.55\textwidth}
    \centering
    \vspace{-9mm}
    \includegraphics[width=\linewidth]{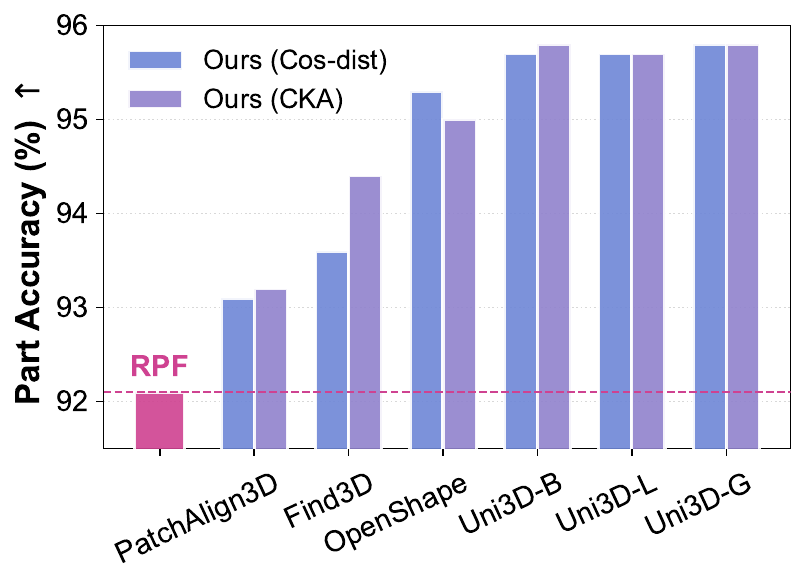}
    \vspace{-7mm}
    \caption{\textbf{Impact of different teachers on distillation.} We compare the Part Accuracy of \method across $\loss_{\text{cos-dist}}$ and $\loss_{\text{CKA}}$ across various 3D foundation models as teachers on Breaking Bad dataset~\cite{sellan2022breaking}. 
    The dashed line indicates the RPF baseline~\cite{sun2025rectified}.}
    \label{fig:teacher}
    \vspace{-6mm}
\end{wrapfigure}

Figure~\ref{fig:teacher} evaluates how the choice of teacher encoder affects downstream assembly when used for alignment. We compare several pretrained 3D foundation models as teachers while keeping the student backbone, training protocol, and alignment formulation fixed, and report Part Accuracy for both token-wise cosine matching ($\loss_{\text{cos-dist}}$) and relational topology alignment ($\loss_{\text{CKA}}$). Overall, stronger teachers consistently translate into better assembly performance. 

Specifically, across teachers, we find that \textbf{Uni3D} yields the most reliable gains under both objectives and achieves the highest Part Accuracy, outperforming PatchAlign3D, Find3D, and OpenShape by a clear margin. Notably, this advantage holds across Uni3D variants, indicating that teacher choice is not merely a matter of model scale but of how well the representation encodes assembly-relevant geometric structure. These results align with our correlation analysis: teachers that better capture geometry/contact cues provide more effective supervision for point-flow assembly.  We refer the readers to the supplementary material for additional visualization that corroborates this.

\vspace{10pt}
\subsection{Emergent Spatial Structure in Later Layers}

In addition to the teacher selection choice, it also remains an important exploration to choose an effective alignment target. To this end, we analyze how spatial structure emerges across layers of the \emph{unaligned} RPF flow backbone. We extract intermediate features from each transformer layer and evaluate four complementary metrics on frozen representations: \textbf{Boundary Contrast}, which measures how sharply features change across inter-part boundaries; \textbf{LDS}, which measures whether features preserve local geometric neighborhoods relative to distant points; \textbf{Part Silhouette}, which measures how well features cluster by part identity; and \textbf{Pose Discrimination}, which measures sensitivity to rigid part motions by comparing features from the assembled configuration to those from a pose-deformed version (per-part rotated/translated). We provide precise definitions and implementation details for all metrics in the supplementary material.
\begin{figure}[t]
    \centering
    \includegraphics[width=0.98\textwidth]{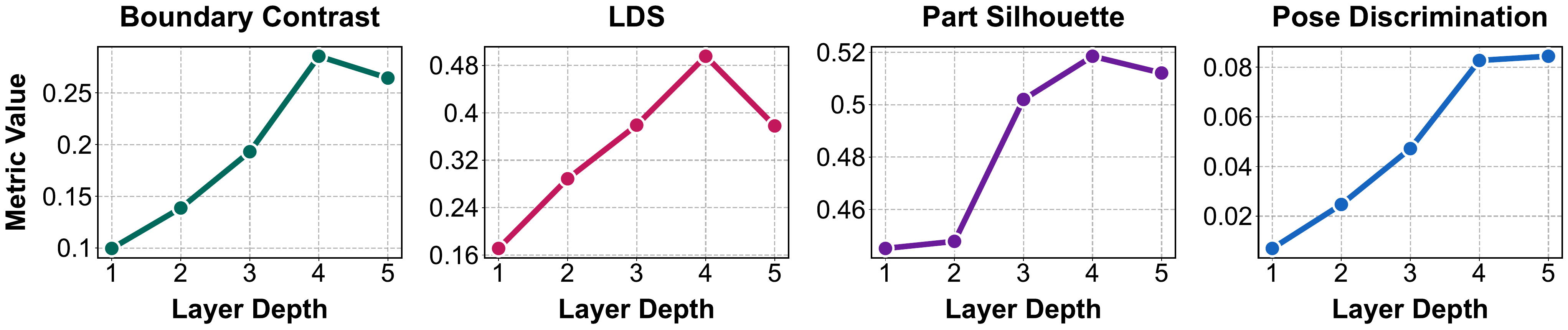}
    \vspace{-1mm}
    \caption{\textbf{Spatial structure emerges in later layers.}
    We measure four spatial metrics across layers of the RPF flow model \emph{without} alignment.
    All metrics increase with depth, indicating that the model progressively resolves spatial part structure in its later layers.
    }\vspace{-10pt}
    \label{fig:geometry_metric}
\end{figure}

As shown in Fig.~\ref{fig:geometry_metric}, all four metrics increase with layer depth, indicating that later layers progressively organize features into assembly-relevant spatial structure. Boundary transitions become sharper, local geometry is more coherently represented, and part-level clusters become more separable. The increase in pose discrimination further suggests that pose-awareness—necessary
\begin{wrapfigure}{r}{0.44\textwidth}
    \centering
    \vspace{-8mm}
    \includegraphics[width=\linewidth]{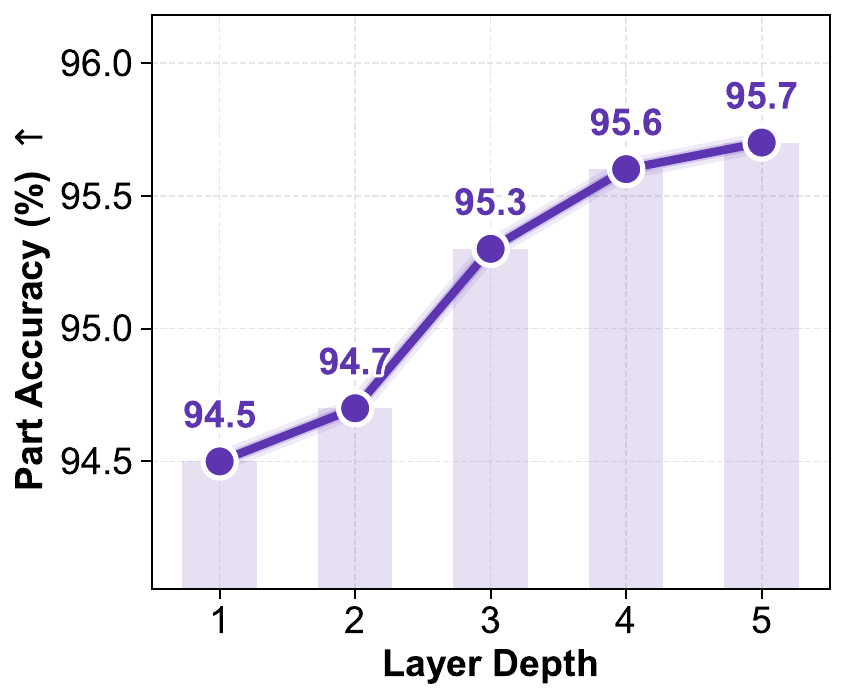}
    \vspace{-6mm}
    \caption{\textbf{Ablation study on alignment layer depth.} Part Accuracy consistently improves when applying alignment to later layers.}
    \label{fig:layer_ablation}
\vspace{-8mm}
\end{wrapfigure}
for 6-DoF recovery—emerges predominantly in later layers where global context is integrated.Motivated by these trends, we apply alignment at late representations, where the model is actively forming the global structure and interactions that govern mating.
Figure~\ref{fig:layer_ablation} corroborates this choice: aligning at deeper layers consistently improves Part Accuracy, with the best performance achieved at $l^*{=}5$, outperforming both early-layer alignment and the unaligned baseline. Based on these analyses, in the following, we evaluate the effectiveness of the proposed framework and compare it against existing methods.

\section{Experimental Results}
\label{sec:experiments}

\subsection{Experimental Setup}
\label{sec:dataset}

\noindent\textbf{Datasets.} 
We evaluate our method on six benchmark datasets, such as Breaking Bad~\cite{sellan2022breaking}, TwoByTwo~\cite{qi2025two}, PartNet-Assembly~\cite{xu2025spaformer}, Fractura~\cite{li2025garf}, Fantastic Breaks~\cite{lamb2023fantastic}, Breaking Bad-\textit{Artifact}~\cite{sellan2022breaking}, that span diverse assembly complexities, ranging from irregular geometric fractures to semantically meaningful multi-part furniture and functional pairwise interactions. All models are trained across all object categories within each benchmark. We refer the readers to the appendix for more details of the datasets.\vspace{-5pt}

\vspace{2mm}
\medbreak
\noindent\textbf{Evaluation Metrics.} 
Following the rigorous evaluation protocols established in recent literature~\cite{sun2025rectified}, we quantitatively assess assembly performance using three primary metrics:

\begin{itemize}[leftmargin=*,itemsep=6pt,topsep=4pt]
    \item \textit{Part Accuracy (PA):} The primary indicator of overall assembly success. It is defined as the percentage of parts whose Chamfer Distance to their ground-truth assembled positions is strictly below a predefined threshold $\tau=0.01$. 
    \item \textit{Rotation Error (RE):} We measure the geodesic distance (in degrees) between the predicted and ground-truth rotation matrices using the Rodrigues formula. As pointed out by Sun~\etal~\cite{sun2025rectified}, this provides a mathematically proper distance metric on the SO(3) manifold, avoiding the representation singularities inherent to the RMSE of Euler angles used in older benchmarks.
    \item \textit{Translation Error (TE):} The Root Mean Square Error (RMSE) of the predicted translation vectors, measured in centimeters (cm).
\end{itemize}

\newcolumntype{C}{>{\centering\arraybackslash}p{1.8cm}} 

\newcommand{\best}[1]{\cellcolor{blue!20}\textbf{#1}}
\newcommand{\second}[1]{\cellcolor{blue!5}#1}

\begin{table}[!t]
\centering
\caption{\textbf{Quantitative comparison on 3D shape assembly benchmarks.} The \colorbox{blue!20}{\textbf{best}} and \colorbox{blue!5}{second best} results are highlighted. For fair comparison, we re-evaluate all methods under a unified evaluation protocol. See supplementary material for details.}

\vspace{-4mm}

\begin{subtable}{\linewidth}
\centering
\caption{\textbf{Breaking Bad}~\cite{sellan2022breaking}: Multi-part \textit{geometric} shape assembly benchmark. Results are grouped by the number of parts per object: 2 to 20 (left) and 21 to 33 (right).}
\vspace{-3mm}
\resizebox{\linewidth}{!}{
\begin{tabular}{l CCC CCC }
\toprule
& \multicolumn{3}{c}{\textbf{Breaking Bad - Everyday} [2,20]} 
& \multicolumn{3}{c}{\textbf{Breaking Bad - Everyday} [21,33]}  \\

\cmidrule(lr){2-4} \cmidrule(lr){5-7}

\textbf{Methods} 
& PA (\%) $\uparrow$ & RE ($^\circ$) $\downarrow$ & TE (cm) $\downarrow$ 
& PA (\%) $\uparrow$ & RE ($^\circ$) $\downarrow$ & TE (cm) $\downarrow$   \\
\midrule

Jigsaw~\cite{lu2023jigsaw}             & 69.7 & 30.2 & 8.9 & 12.0 & 92.1 & 20.7 \\
PMTR~\cite{lee2024pmtr}                & 70.6 & 24.9 & 14.9 & 8.4 & 72.2 & 20.1 \\
CMNet~\cite{lee2025cmnet}              & 80.4 & 19.7 & 11.6 & 24.7 & 67.7 & 20.8  \\
GARF~\cite{li2025garf}                 & 92.4 & \best{7.4} & 2.7 & 25.1 & 68.0 & 32.9 \\
RPF~\cite{sun2025rectified}            & \second{93.2} & 16.0 & 4.3  & 62.1 & 77.3 & 15.2 \\

\midrule

\textbf{Ours}$_\textrm{NT-Xent}$ & 92.9 & 17.8 & 4.7  & 62.6 & 77.0 & 15.2  \\
\textbf{Ours}$_\textrm{Cos-dist}$ & \best{95.7}  & {9.0} & \second{2.2} & \best{72.4} & \best{64.3} & \best{12.3} \\
\textbf{Ours}$_\textrm{CKA}$ & \best{95.7}  & \second{8.6} & \best{2.1} &  \second{71.7} & \second{64.8} & \second{12.5}  \\
\bottomrule
\end{tabular}
}
\end{subtable}

\vspace{3mm}

\begin{subtable}{\linewidth}
\centering
\caption{\textbf{PartNet-Assembly}~\cite{xu2025spaformer}: Multi-part \textit{semantic} shape assembly, and \textbf{TwoByTwo}~\cite{qi2025two}: \textit{inter-object} assembly benchmark.}
\vspace{-3mm}
\resizebox{\linewidth}{!}{
\begin{tabular}{l CCC CCC }
\toprule
& \multicolumn{3}{c}{\textbf{PartNet-Assembly}} 
& \multicolumn{3}{c}{\textbf{TwoByTwo}}  \\

\cmidrule(lr){2-4} \cmidrule(lr){5-7}

\textbf{Methods} 
& PA (\%) $\uparrow$ & RE ($^\circ$) $\downarrow$ & TE (cm) $\downarrow$ 
& PA (\%) $\uparrow$ & RE ($^\circ$) $\downarrow$ & TE (cm) $\downarrow$  \\
\midrule

RPF~\cite{sun2025rectified} & 59.8 & 46.2 & 21.5  & 65.4 & 15.8 & 11.9 \\
\midrule

\textbf{Ours}$_\textrm{NT-Xent}$ & 65.4 & 43.6 & 20.2  & 60.4 & 17.2 & 12.8 \\
\textbf{Ours}$_\textrm{Cos-dist}$ & \second{67.8} & \second{41.6} & \second{19.1} & \second{68.9} & \second{12.0} & \second{9.5} \\
\textbf{Ours}$_\textrm{CKA}$ & \best{69.1} & \best{40.8} & \best{18.8} & \best{71.5} & \best{10.0} & \best{7.6} \\

\bottomrule
\end{tabular}
}
\end{subtable}

\label{tab:main}
\end{table}

\subsection{Implementation Details}
\label{sec:implementation}

We implement our method in PyTorch Lightning~\cite{falcon2019lightning} and train on 8 NVIDIA GH200 GPUs with a total batch size of 256 for 2,000 epochs.
We use AdamW~\cite{loshchilovdecoupled} with a learning rate of $5 \times 10^{-4}$, halved every 200 epochs after 1,000 epochs.
For the overlap-aware encoder, we use the official pretrained checkpoint from RPF~\cite{sun2025rectified}.
Unless otherwise stated, all experiments use Uni3D-L~\cite{zhou2023uni3d} as the teacher encoder, CKA loss ($\loss_{\text{CKA}}$) as the alignment objective with $n{=}1\text{,}024$ subsampled tokens. 
The alignment weight is set to $\lambda{=}0.5$, and for the NT-Xent variant, we use a temperature of $\tau{=}0.07$.

\subsection{Experimental Results}
\noindent\textbf{Multi-part Assembly.} 
Table~\ref{tab:main} summarizes results on three complementary multi-part assembly regimes: \textit{geometric} reassembly (Breaking Bad), \textit{semantic} part assembly (PartNet-Assembly), and \textit{inter-object} assembly under stronger distribution shift (TwoByTwo). 
We report Part Accuracy together with rotation and translation errors. Across all benchmarks, casting training as teacher--student distillation consistently improves the strong RPF baseline, confirming that injecting pretrained geometric priors into the flow backbone is broadly beneficial for 3D assembly.

\vspace{10mm}
\begin{figure}[!t]
    \centering
    \includegraphics[width=0.98\textwidth]{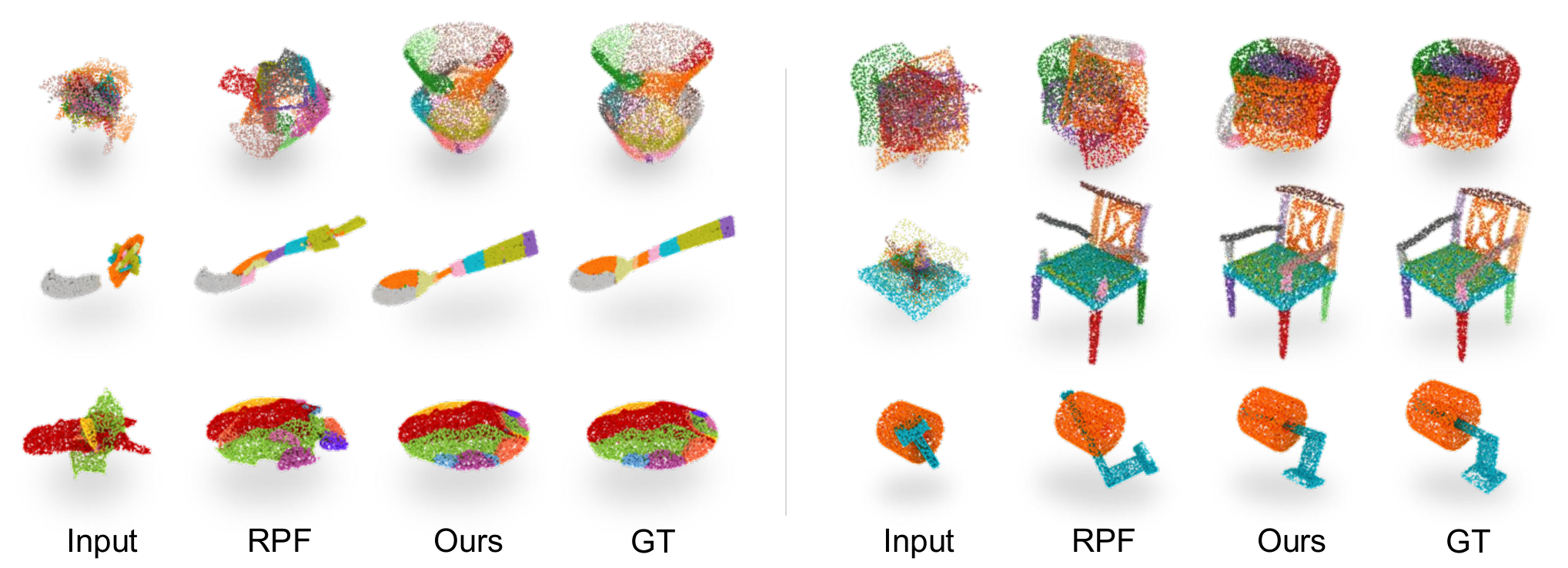}
    \vspace{-1mm}
    \caption{\textbf{Qualitative Comparison.} While the baseline RPF often struggles with precise part positioning and fails to resolve complex inter-part relations, ours consistently produces structurally coherent assemblies that closely match the ground truth.
    }
    \label{fig:qual}
\end{figure} 
On Breaking Bad (2 to 20 parts), even our simplest instantiation---token-wise cosine alignment---already yields substantial gains over RPF in Part Accuracy and pose errors, demonstrating that representation alignment transfers effectively to the 3D point-flow setting. Topology alignment (CKA) matches or further improves these results, achieving the best overall pose accuracy. The many-part setting (21 to 33 parts) highlights the scalability challenges faced by existing methods: correspondence-based approaches such as Jigsaw, CMNet, and PMTR degrade significantly as the combinatorial complexity of multi-fragment matching grows, while GARF, despite being competitive on the few-part split, drops sharply to lower Part Accuracy. 
In contrast, RPF maintains reasonable performance in this regime, and our alignment further amplifies its scalability, delivering massive gains in both Part Accuracy and pose errors. Figure~\ref{fig:qual} provides qualitative examples where RPF struggles with precise part positioning, while our method produces assemblies closely matching the ground truth.

On PartNet-Assembly and TwoByTwo, CKA delivers the greatest improvements, suggesting that explicitly preserving relational structure is particularly valuable when assembly involves structured part interactions and domain shift. We also note that NT-Xent degrades performance on TwoByTwo, indicating that enforcing per-point discriminability can be counterproductive in inter-object settings where the teacher already separates distinct objects (see supplementary).

\medbreak
\noindent\textbf{Loss combination ablation.}
The comparisons above evaluate the three alignment objectives individually, where $\mathcal{L}_{\text{CKA}}$ consistently performs best. A natural follow-up is whether \emph{combining} them provides complementary gains over $\mathcal{L}_{\text{CKA}}$ alone. We study this on TwoByTwo, as its inter-object setting most strongly exposes the differences among objectives (Tab.~\ref{tab:main}), making it the most discriminative regime for isolating each loss's contribution.

\setlength{\columnsep}{4mm}
\begin{wraptable}{r}{0.35\linewidth}
\centering
\vspace{-8mm}
\resizebox{\linewidth}{!}{%
\begin{tabular}{lc}
\toprule
$\mathcal{L}_{\text{align}}$ & PA $\uparrow$ \\
\midrule
\rowcolor{gray!18}
$\mathcal{L}_{\text{CKA}}$ \textbf{(Ours)} & \textbf{71.5} \\
$\mathcal{L}_{\text{CKA}} + \mathcal{L}_{\text{cos-dist}}$ & 70.0 \\
$\mathcal{L}_{\text{CKA}} + \mathcal{L}_{\text{NT-Xent}}$ & 68.5 \\
$\mathcal{L}_{\text{CKA}} + \mathcal{L}_{\text{cos-dist}} + \mathcal{L}_{\text{NT-Xent}}$ & 67.7 \\
\bottomrule
\end{tabular}%
}
\vspace{-3mm}
\caption{Loss combination ablation on TwoByTwo.}
\label{tab:supp_loss_combo}
\vspace{-6mm}
\end{wraptable}
Table~\ref{tab:supp_loss_combo} reports Part Accuracy on TwoByTwo when augmenting $\mathcal{L}_{\text{CKA}}$ with token-wise objectives. In all cases, combining multiple losses degrades performance, with the full combination performing worst. This indicates that $\mathcal{L}_{\text{CKA}}$'s relational signal already captures the relevant structural information, and token-wise objectives introduce competing gradients that dilute the alignment.

\begin{table}[t]
\centering\label{zero}
\caption{\textbf{Zero-shot Evaluation on Breaking Bad - Artifact~\cite{sellan2022breaking}, FRACTURA~\cite{li2025garf} and Fantastic Breaks~\cite{lamb2023fantastic} datasets}.
} 

\resizebox{\linewidth}{!}{
\begin{tabular}{l CCC CCC CCC }
\toprule
& \multicolumn{3}{c}{\textbf{Breaking Bad - Artifact}}
& \multicolumn{3}{c}{\textbf{FRACTURA}}
& \multicolumn{3}{c}{\textbf{Fantastic Breaks}}\\

\cmidrule(lr){2-4} \cmidrule(lr){5-7} \cmidrule(lr){8-10}

\textbf{Methods} 
& PA (\%) $\uparrow$ & RE ($^\circ$) $\downarrow$ & TE (cm) $\downarrow$ 
& PA (\%) $\uparrow$ & RE ($^\circ$) $\downarrow$ & TE (cm) $\downarrow$ 
& PA (\%) $\uparrow$ & RE ($^\circ$) $\downarrow$ & TE (cm) $\downarrow$ \\
\midrule

GARF~\cite{li2025garf} & 91.4 & \second{8.7} & 3.0 & 44.2 & 37.8 & 26.9 & 88.3 & 8.2 & 3.0 \\
RPF~\cite{sun2025rectified} & 88.3 & 20.9 & 5.3 & 68.1  & 50.1 & 11.2 & 96.9 & 6.3 & 1.5 \\

\midrule

\textbf{Ours}$_\textrm{NT-Xent}$ & 87.0 & 22.9 & 5.9 & 71.7 & 48.9 & 10.6 & \second{97.6} & 6.3 & 1.5 \\
\textbf{Ours}$_\textrm{Cos-dist}$ & \second{93.2} & 11.3 & \second{2.8} & \second{74.9} & \best{35.5} & \second{7.9} & \best{97.7} & \second{4.5} & \second{1.1} \\
\textbf{Ours}$_\textrm{CKA}$ & \best{94.4} & \best{8.0} & \best{2.1} & \best{76.0} & \second{36.4} & \best{7.7} & 97.2 & \best{3.5} & \best{0.9} \\

\bottomrule

\end{tabular}
}
\label{tab:zeroshot}
\end{table}

\medbreak
\noindent\textbf{Zero-shot Transfer to Unseen Datasets.}
To assess robustness under domain shift, we evaluate models trained on the Breaking Bad everyday split in a zero-shot manner on three unseen datasets: Breaking Bad artifact split~\cite{sellan2022breaking} (synthetic, unseen object categories), FRACTURA~\cite{li2025garf} (mixed synthetic and real fractures across scientific domains), and Fantastic Breaks~\cite{lamb2023fantastic} (real-world scanned objects). Table~\ref{tab:zeroshot} summarizes the results without any fine-tuning. Our alignment transfers substantially better than the RPF baseline across all three datasets. In particular, ours with a CKA objective achieves the best overall performance, attaining the lowest pose errors on both FRACTURA and Fantastic Breaks. Token-wise cosine alignment consistently improves over RPF, whereas the contrastive NT-Xent objective is less reliable under shift, often degrading pose accuracy. Overall, these results confirm that topological alignment yields strong generalization to unseen distributions, supporting our claim that transferring relational structure is particularly beneficial for 3D assembly.

\begin{figure}[t]
    \centering
    \includegraphics[width=0.98\textwidth]{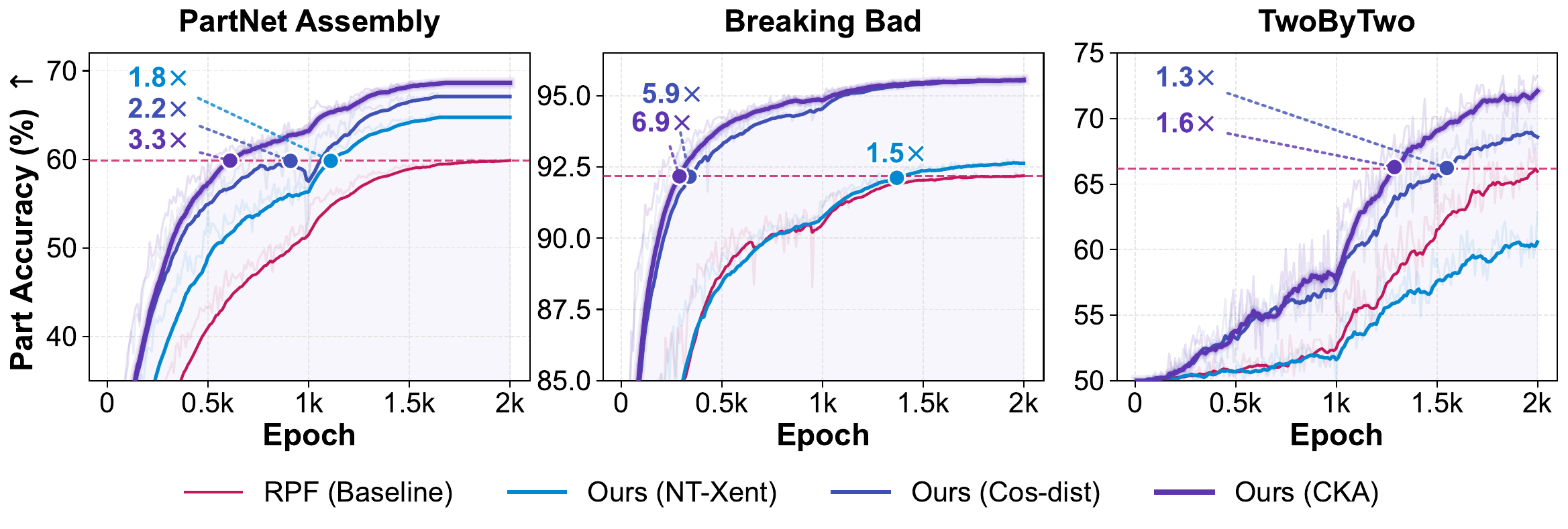}
    \vspace{-3mm}
    \caption{\textbf{Convergence comparison.} We monitor the validation Part Accuracy of Ours (CKA) (\textcolor[HTML]{5E35B1}{\rule[.5ex]{1.2em}{2pt}}) against the RPF (\textcolor[HTML]{C2185B}{\rule[.5ex]{1.2em}{1.5pt}}) and other alignment strategies including Ours (NT-Xent) (\textcolor[HTML]{0288D1}{\rule[.5ex]{1.2em}{1.5pt}}) and Ours (Cos-dist) (\textcolor[HTML]{3F51B5}{\rule[.5ex]{1.2em}{1.5pt}}) over training epochs across three datasets. The dashed horizontal line (\textcolor[HTML]{C2185B}{\rule[.5ex]{0.4em}{1.5pt}\hspace{1.5pt}\rule[.5ex]{0.4em}{1.5pt}}) represents the peak accuracy of the baseline. The annotated multipliers indicate the convergence speedup relative to the baseline to reach its peak performance.}\vspace{-10pt}
    \label{fig:convergence}
\end{figure}
\medbreak
\noindent\textbf{Convergence Analysis.}
Figure~\ref{fig:convergence} compares validation Part Accuracy as training progresses on PartNet-Assembly, Breaking Bad, and TwoByTwo. Across all datasets, alignment accelerates optimization relative to the RPF baseline, and the proposed topology alignment provides the most consistent speedup. On {Breaking Bad}, TORA reaches the baseline’s peak performance substantially earlier (about $6.9\times$ faster), and also converges to a higher final accuracy; token-wise cosine alignment also speeds up training but with a smaller gain. On {PartNet-Assembly}, TORA again yields the largest acceleration, reaching the baseline peak $3.3\times$ sooner, compared to $2.2\times$ for cosine and $1.8\times$ for NT-Xent. The effect is most pronounced under domain shift on {TwoByTwo}, where TORA improves both convergence and final accuracy, while token-wise alignment provides more limited acceleration and NT-Xent remains less reliable. Overall, these results indicate that explicitly matching relational structure provides a richer and more task-aligned training signal, enabling faster and more stable optimization across assembly regimes.

\begin{wraptable}{r}{0.55\textwidth}
\centering
\vspace{-4mm}
\caption{\textbf{Per-step Training Overhead.} Measured on a single NVIDIA GH200 GPU with a batch size of 1. Statistics are averaged following a 50-step warmup.}
\label{tab:efficiency}
\vspace{2mm}
\resizebox{\linewidth}{!}{
\begin{tabular}{ccCCc}
\toprule
\multirow{2}{*}{\textbf{Methods}} & \multirow{2}{*}{$\mathcal{L}_{\text{align}}$} & \textbf{Memory} & \textbf{Time} & \textbf{Throughput} \\
 & & (GB) & (ms) & (steps/s) \\
\midrule
\multirow{3}{*}{\begin{tabular}[c]{@{}c@{}}\textbf{Ours} \\ (Online Teacher)\end{tabular}} 
 & NT-Xent  & 4.79 & 124.85 & 8.01 \\
 & cos-dist & 4.85 & 122.02 & 8.20 \\
 & CKA      & 4.72 & 128.39 & 7.79 \\
\midrule
\multirow{3}{*}{\begin{tabular}[c]{@{}c@{}}\textbf{Ours} \\ (Offline Teacher)\end{tabular}} 
 & NT-Xent  & 3.48 & 104.19 & 9.60 \\
 & cos-dist & 3.48 & 104.69 & 9.55 \\
 & CKA      & 3.48 & 104.22 & 9.59 \\
\midrule
RPF~\cite{sun2025rectified} & - & 3.30 & 101.40 & 9.86 \\
\bottomrule
\end{tabular}
}
\vspace{-5mm}
\end{wraptable}
\medbreak
\noindent\textbf{Efficiency Analysis.} We profile the training overhead of TORA against the baseline RPF in Tab.~\ref{tab:efficiency}. 
Crucially, because the teacher encoder remains completely frozen, its representations can be practically precomputed and cached \textit{offline}. 
Under this standard feature-caching setting, TORA incurs near-zero practical overhead, requiring only an additional 0.18 GB peak VRAM (+5.5\%) and $\sim$3\,ms per step (+2.8\%) compared to the baseline—negligible on modern hardware. However, when executing an \textit{online} forward pass, the computational footprint is shown to increase. Nevertheless, it remains highly manageable, as only approximately 1.4 GB and 27 ms are additionally induced. We highlight that this is achieved via our implementation to efficiently compute Gram matrices through random subsampling, as discussed in Sec.~\ref{sec:architecture}. Ultimately, this confirms that TORA is highly scalable, preserving the massive convergence speedup without compromising training throughput, while adding strictly zero overhead during inference.

\vspace{-2mm}
\section{Conclusion}
In this work, we have introduced \method, a topology-first teacher–student alignment framework for robust 3D point-flow assembly. By distilling inter-point relational structure from a frozen pretrained 3D encoder into a flow-matching assembly model, \method injects interaction-aware geometric priors while preserving the original inference pipeline and incurring zero test-time overhead. Extensive experiments across semantic, geometric, and inter-object assembly benchmarks demonstrate consistent improvements over strong flow-based baselines, with particularly pronounced gains under domain shift, where relational topology transfer is most beneficial. Our analysis further clarifies what makes an effective teacher for 3D assembly—geometry- and contact-centric signals rather than category semantics—and shows that topology alignment accelerates convergence and improves final accuracy. We hope these findings encourage broader use of relational distillation objectives for 3D generative transport models and enable more robust assembly in real-world robotics and graphics applications.

\section*{Acknowledgements}
This research was supported by the ETH AI Center through an ETH AI Center postdoctoral fellowship to Sunghwan Hong, Swiss AI Initiative (a144), and ELLIOT (Grant Agreement 101214398).

\appendix
\renewcommand{\thefigure}{A\arabic{figure}}
\renewcommand{\thetable}{A\arabic{table}}
\setcounter{figure}{0}
\setcounter{table}{0}

\begin{center}
    \Large \textbf{Supplementary Material}
\end{center}
\vspace{1em}

\noindent
In this supplementary material, we present additional information and analyses not included in the main paper. The contents are organized as follows:

\begin{itemize}[leftmargin=1.5em, itemsep=1pt, topsep=2pt]
    \item \textbf{Section~\ref{sec:supp_teacher_probes}}: Probing protocols for evaluating teacher representations.
    \item \textbf{Section~\ref{sec:supp_layer_metrics}}: Metrics for measuring emergent spatial structure across layers.
    \item \textbf{Section~\ref{sec:supp_teacher_details}}: Teacher encoder specifications.
    \item \textbf{Section~\ref{sec:supp_additional_analysis}}: Extended teacher selection analyses.
    \item \textbf{Section~\ref{sec:supp_implementation}}: Additional implementation details.
    \item \textbf{Section~\ref{sec:supp_eval}}: Evaluation protocol clarification.
    \item \textbf{Section~\ref{sec:supp_validation}}: Validation of experimental results.
    \item \textbf{Section~\ref{sec:suppl_failure}}: Failure cases and future directions.
    \item \textbf{Section~\ref{sec:additional_qual}}: Additional qualitative results.
\end{itemize}

\section{Probing Teacher Representations}
\label{sec:supp_teacher_probes}

Here, we provide the details of the probing analyses used to contextualize the teacher-selection results in Sec.~\mainsec{4.1} (Fig.~\mainsec{4}). Each frozen teacher encoder is evaluated with four complementary probes: two task-based probes that involve training a lightweight linear head, and two spatial probes computed directly from the frozen feature representations. We study six pretrained 3D encoders spanning a range of training objectives and model scales; details of the encoders are provided in Sec.~\ref{sec:supp_teacher_details}. All probes are performed on the Breaking Bad-Everyday~\cite{sellan2022breaking} dataset using the same train/test split as in the shape assembly benchmark.

\subsection{Task-based Probes}
\label{sec:supp_task_probes}

\medbreak\noindent\textbf{Object Classification.}
This probe measures the extent to which the teacher representation captures global semantic information. We first apply global average pooling to the per-point features produced by the frozen 3D encoder to obtain a single shape-level descriptor, and then train a linear classifier to predict the object category among the 20 classes in Breaking Bad-Everyday. The classifier is trained for 10 epochs using cross-entropy loss and AdamW~\cite{loshchilovdecoupled}, with a learning rate of $10^{-3}$ and a batch size of 8. We report top-1 classification accuracy on the held-out test set. This probe corresponds to Fig.~\mainsec{4}(a).

\medbreak\noindent\textbf{Mating-Surface Segmentation.}
This probe evaluates whether the teacher features encode contact-aware geometry and part-to-part interaction cues. We formulate mating-surface prediction as a per-point binary classification task on the frozen point-wise features. Following Sun~\etal~\cite{sun2025rectified}, a point is labeled as mating (positive) if it has at least one neighbor from a different part within an adaptive overlap threshold $\tau = \sqrt{2A/N}$, where $A$ denotes the total surface area and $N=5\text{,}000$ is the number of sampled points. This threshold approximates the expected nearest-neighbor distance on the surface, so that a point is labeled as mating whenever a point from another part lies within this range. The probe is trained for 10 epochs using binary cross-entropy loss and AdamW, with a learning rate of $10^{-3}$ and a batch size of 8. We report the F1 score for the positive (mating) class. This probe corresponds to Fig.~\mainsec{4}(b).

\subsection{Spatial Probes}
\label{sec:supp_spatial_probes}

The following two metrics are computed directly from frozen per-point features, without training any additional prediction head. Let $\tilde{\mathbf{h}}_i = \mathbf{h}_i / \|\mathbf{h}_i\|_2$ denote the $\ell_2$-normalized feature of point $i$, $\mathcal{N}_k(i)$ the set of $k$ nearest neighbors of point $i$ in Euclidean coordinate space, and $p(i)$ the part label of point $i$. We use $k{=}6$ throughout.

\medbreak\noindent\textbf{Local-vs-Distant Similarity (LDS).}
LDS~\cite{huang1997image,singh2025irepa} measures the extent to which features of spatially nearby points are more similar than those of distant points. We define local similarity as the mean cosine similarity between each point and its $k$ nearest neighbors, and distant similarity as the mean cosine similarity over all point pairs whose Euclidean distance exceeds a threshold $d_\mathrm{far}$. We set $d_\mathrm{far}$ to the 75th percentile of all pairwise Euclidean distances within each sample, allowing the threshold to adapt to the spatial extent of the object:
\begin{equation}
    \mathrm{LDS}(\tilde{\mathbf{h}}, \mathbf{x}) = \frac{1}{N}\sum_{i=1}^{N}\frac{1}{k}\sum_{j\in \mathcal{N}_k(i)} \mathrm{sim}(\tilde{\mathbf{h}}_i, \tilde{\mathbf{h}}_j) - \frac{1}{|\mathcal{F}|}\sum_{(i,j)\in \mathcal{F}} \mathrm{sim}(\tilde{\mathbf{h}}_i, \tilde{\mathbf{h}}_j),
\end{equation}
where $\mathcal{F} = \{(i,j) : \|\mathbf{x}_i - \mathbf{x}_j\|_2 \geq d_\mathrm{far},\, i \neq j\}$ denotes the set of distant pairs. Higher values indicate stronger spatial locality in feature space. This probe corresponds to Fig.~\mainsec{4}(c).

\medbreak\noindent\textbf{Part Silhouette.}
We compute the silhouette score~\cite{rousseeuw1987silhouettes} using cosine distance, defined as $d(i,j) = 1 - \mathrm{sim}(\tilde{\mathbf{h}}_i, \tilde{\mathbf{h}}_j)$, and treat the ground-truth part labels as cluster assignments. For each point $i$ belonging to part $p(i)$, we define the intra-part distance $a(i)$ and the nearest inter-part distance $b(i)$ as:
\begin{align}
    a(i) = \frac{1}{|C_{p(i)}|-1}\sum_{\substack{j \in C_{p(i)} \\ j \neq i}} d(i,j), \ \ \ \ 
    b(i) = \min_{q \neq p(i)} \frac{1}{|C_q|}\sum_{j \in C_q} d(i,j),
\end{align}
where $C_p = \{j : p(j) = p\}$ denotes the set of points belonging to part $p$. The Part Silhouette score is then defined as:
\begin{equation}
    \mathrm{PS}(\tilde{\mathbf{h}}, p) = \frac{1}{N}\sum_{i=1}^{N} \frac{b(i) - a(i)}{\max(a(i),\, b(i))}.
\end{equation}
The score lies in $[-1, 1]$, where values close to $1$ indicate well-separated parts, while negative values suggest that the feature representation does not respect part boundaries. This probe corresponds to Fig.~\mainsec{4}(d).
\section{Measuring Emergent Spatial Structure}
\label{sec:supp_layer_metrics}

To motivate the alignment-layer choice in Sec.~\mainsec{4.3} (Fig.~\mainsec{6}), we analyze how spatial structure emerges across layers of the \emph{unaligned} RPF flow backbone. Specifically, we extract intermediate features from each transformer layer and evaluate four geometry-related metrics on the resulting frozen representations. All metrics are computed on $\ell_2$-normalized intermediate features, $\tilde{\mathbf{h}}^{(l)}_i = \mathbf{h}^{(l)}_i / \|\mathbf{h}^{(l)}_i\|_2$, extracted from transformer layer $l$. We reuse the notation $\mathcal{N}_k(i)$, $p(i)$, and $k{=}6$ from Sec.~\ref{sec:supp_spatial_probes}.

\medbreak\noindent\textbf{Boundary Contrast.}
This metric measures how sharply features change across inter-part boundaries. For each point $i$, we partition its spatial neighbors into interior pairs, $\mathcal{I}(i) = \{j \in \mathcal{N}_k(i) : p(j) = p(i)\}$, and boundary pairs, $\mathcal{B}(i) = \{j \in \mathcal{N}_k(i) : p(j) \neq p(i)\}$. Boundary Contrast is defined as the difference between the mean similarity of interior pairs and that of boundary pairs:
\begin{equation}
    \mathrm{BC}(\tilde{\mathbf{h}}^{(l)}, \mathbf{x}, p) = \frac{1}{|\mathcal{I}|}\sum_{(i,j)\in \mathcal{I}} \mathrm{sim}(\tilde{\mathbf{h}}^{(l)}_i, \tilde{\mathbf{h}}^{(l)}_j) - \frac{1}{|\mathcal{B}|}\sum_{(i,j)\in \mathcal{B}} \mathrm{sim}(\tilde{\mathbf{h}}^{(l)}_i, \tilde{\mathbf{h}}^{(l)}_j),
\end{equation}
where $\mathcal{I} = \bigcup_i \mathcal{I}(i)$ and $\mathcal{B} = \bigcup_i \mathcal{B}(i)$ denote the sets aggregated over all points. A value of zero indicates that features are equally similar within and across part boundaries, whereas higher values indicate sharper feature transitions at boundaries. This metric corresponds to Fig.~\mainsec{6} (leftmost).

\medbreak\noindent\textbf{Local-vs-Distant Similarity (LDS).}
This metric is defined in Sec.~\ref{sec:supp_spatial_probes} and is applied here to the intermediate features $\tilde{\mathbf{h}}^{(l)}$ at each layer. It corresponds to Fig.~\mainsec{6} (second from left).

\medbreak\noindent\textbf{Part Silhouette.}
This metric is defined in Sec.~\ref{sec:supp_spatial_probes} and is likewise applied to the intermediate features $\tilde{\mathbf{h}}^{(l)}$ at each layer. It corresponds to Fig.~\mainsec{6} (second from right).

\medbreak\noindent\textbf{Pose Discrimination.}
This metric measures the sensitivity of features to rigid part transformations. We compute features from two configurations of the same object: the ground-truth assembled configuration and a deformed configuration in which each part is independently rotated and translated. Let $\tilde{\mathbf{h}}^{(l)}_i$ and $\tilde{\mathbf{h}}^{(l)\prime}_i$ denote the normalized features extracted from the assembled and deformed configurations, respectively. Pose Discrimination is defined as:
\begin{equation}
    \mathrm{PD}(\tilde{\mathbf{h}}^{(l)}, \tilde{\mathbf{h}}^{(l)\prime}) = 1 - \frac{1}{N}\sum_{i=1}^{N} \mathrm{sim}(\tilde{\mathbf{h}}^{(l)}_i,\, \tilde{\mathbf{h}}^{(l)\prime}_i).
\end{equation}
A value of zero indicates invariance to the applied perturbation, whereas higher values indicate greater pose sensitivity, which is important for accurate 6-DoF pose recovery. This metric corresponds to Fig.~\mainsec{6} (rightmost).

\begin{table}[t]
\centering
\caption{Overview of pretrained 3D encoders used as teacher models.}
\label{tab:teacher_details}
\vspace{-2mm}
\resizebox{\linewidth}{!}{%
\begin{tabular}{lcccc}
\toprule
\textbf{Model} & \textbf{Backbone} & \textbf{\#Params} & \textbf{Training Data} & \textbf{Training Objective} \\
\midrule
PatchAlign3D~\cite{hadgi2026patchalign3d} & PointBERT~\cite{yu2022pointbert} & 23.5M & \makecell{Objaverse~\cite{deitke2023objaverse}\\(${\sim}$800K)} & \makecell{DINOv2 distill.\\+ CLIP text contr.} \\[3pt]
\midrule
Find3D~\cite{ma2025find} & PTv3~\cite{wu2024ptv3} & 47.1M & \makecell{Objaverse~\cite{deitke2023objaverse}\\(${\sim}$30K)} & \makecell{SigLIP\\contrastive} \\
\midrule
OpenShape~\cite{liu2023openshape} & PointBERT~\cite{yu2022pointbert} & 33.5M & \multirow{4}{*}{\makecell{Objaverse~\cite{deitke2023objaverse},\\ShapeNet~\cite{chang2015shapenet},\\3D-FUTURE~\cite{fu20213d},\\ ABO~\cite{collins2022abo} (${\sim}$876K)}} & \multirow{4}{*}{\makecell{Text-image-3D\\contrastive}} \\
\cmidrule(lr){1-3}
Uni3D-B~\cite{zhou2023uni3d} & \multirow{3}{*}{ViT~\cite{dosovitskiy2020vit}} & 86M & & \\
Uni3D-L~\cite{zhou2023uni3d} & & 303M & & \\
Uni3D-G~\cite{zhou2023uni3d} & & 1B & & \\
\bottomrule
\end{tabular}%
}
\end{table}

\section{Teacher Encoder Details}
\label{sec:supp_teacher_details}

Table~\ref{tab:teacher_details} summarizes the pretrained 3D encoders used as teachers in our alignment framework (Sec.~\mainsec{4.1}--\mainsec{4.2}). We select these models to span diverse training objectives, feature granularities, and model scales, enabling a systematic analysis of which teacher properties are most beneficial for 3D assembly.

\medbreak\noindent\textbf{PatchAlign3D}~\cite{hadgi2026patchalign3d} is a two-stage encoder that first distills dense DINOv2 features~\cite{oquab2023dinov2} into 3D patch tokens, and then aligns them with CLIP text embeddings~\cite{radford2021learning} using a multi-positive contrastive objective. Trained on ${\sim}800$K Objaverse shapes~\cite{deitke2023objaverse}, it produces local patch-level features and achieves strong zero-shot segmentation without requiring multi-view rendering at inference time.

\medbreak\noindent\textbf{Find3D}~\cite{ma2025find} is an open-world part segmentation model that produces per-point features aligned with the SigLIP embedding space~\cite{zhai2023siglip}. It is trained on ${\sim}30$K Objaverse shapes annotated by a data engine built on SAM and Gemini, using a contrastive objective over part-level text embeddings.

\medbreak\noindent\textbf{OpenShape}~\cite{liu2023openshape} learns a joint representation over text, images, and 3D point clouds via tri-modal contrastive learning. It is trained on a large-scale mixture of four 3D datasets (${\sim}876$K shapes) with enriched text descriptions, and produces a global shape embedding aligned with the CLIP embedding space.

\medbreak\noindent\textbf{Uni3D}~\cite{zhou2023uni3d} adopts a vanilla Vision Transformer~\cite{dosovitskiy2020vit} as its 3D backbone, initialized from pretrained 2D ViT weights, and aligns 3D point cloud features with image-text features from a frozen CLIP teacher via contrastive learning. We evaluate three model scales: Uni3D-B (86M), Uni3D-L (303M), and Uni3D-G (1B). All are trained on the same ${\sim}876$K-shape dataset mixture, allowing us to isolate the effect of teacher scale on distillation quality.

\medbreak\noindent\textbf{Point-wise feature propagation.}
Our token-wise alignment and linear probing tasks require dense point-level features, so we propagate each teacher's coarse outputs to the full-resolution input point cloud of size $N$. For models that natively produce patch-level tokens (Uni3D, OpenShape, and PatchAlign3D) or subsampled point features (Find3D), we use inverse distance weighting (IDW) interpolation, following PointNet++~\cite{qi2017pointnet++}. Concretely, for each of the $N$ input points, we identify its $k{=}3$ nearest patch centers or subsampled points in Euclidean space, and compute the dense feature as the distance-weighted average of their features.

\section{Additional Analysis on Teacher Selection}
\label{sec:supp_additional_analysis}
In Sec.~\mainsec{4.1}--\mainsec{4.2} of the main paper, we established that geometry- and contact-centric teacher properties tend to be more predictive of downstream assembly performance than global semantics, and selected Uni3D as our default teacher. Here, we provide additional visualizations and analyses that supplement this study. In addition to the object-centric encoders studied in the main paper (Uni3D, OpenShape, Find3D, PatchAlign3D), we also consider two scene-centric 3D encoders, \textit{Sonata}~\cite{wu2025sonata} and \textit{Concerto}~\cite{zhang2025concerto}, for completeness.

\subsection{Representation Probing}

\begin{wrapfigure}{r}{0.6\textwidth}
    \centering
    \vspace{-11mm}
    \includegraphics[width=\linewidth]{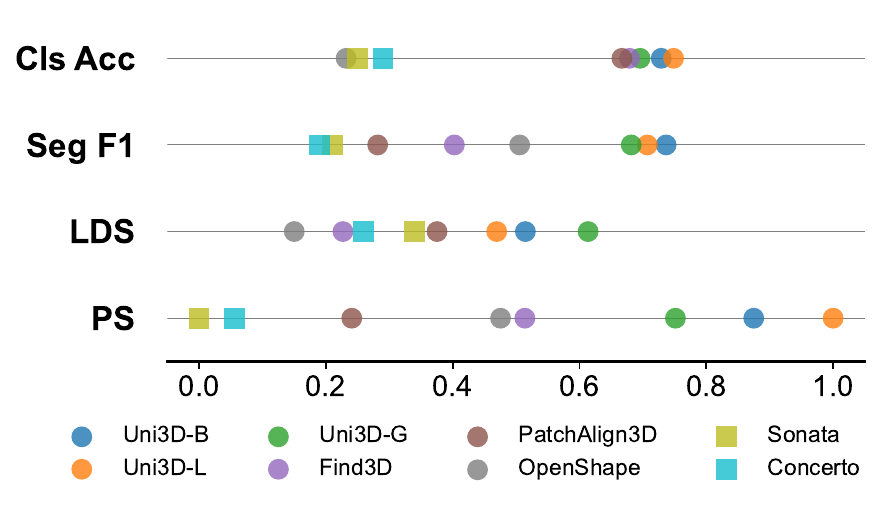}
    \vspace{-6mm}
    \caption{\textbf{Probe metrics across teachers.} We report the four representation probes for all evaluated teachers, extending the analysis to include scene-centric encoders (Sonata, Concerto). Circles denote object-centric encoders and squares denote scene-centric encoders. Part Silhouette (PS) values are min-max normalized for visual clarity.}
    \label{fig:teacher_analysis}
\vspace{-8mm}
\end{wrapfigure}

We apply the four probe metrics introduced in Sec.~\ref{sec:supp_teacher_probes} to all evaluated teachers, including the two scene-centric encoders not covered in the main paper. As shown in Fig.~\ref{fig:teacher_analysis}, the Uni3D family achieves consistently high scores on the geometry- and contact-oriented probes (Seg. F1, LDS and PS), with all three variants ranking among the top across all metrics. This aligns with the correlation trends observed in Fig.~\mainsec{4}, where these probes were most predictive of downstream assembly quality. The OpenShape, Find3D, and PatchAlign3D teachers show more varied profiles across the four probes. However, in contrast, the two scene-centric encoders, Sonata and Concerto, generally cluster toward the lower end (leftmost) on all probes. We hypothesize that this may be due to the discrepancies in data distribution which the models were trained on, \textit{i.e.,} scene-level and object-level, exhibit largely different granularity, and this may have caused detrimental effects on the performance on shape assembly.

\begin{figure}[t]
    \centering
    \includegraphics[width=0.98\textwidth]{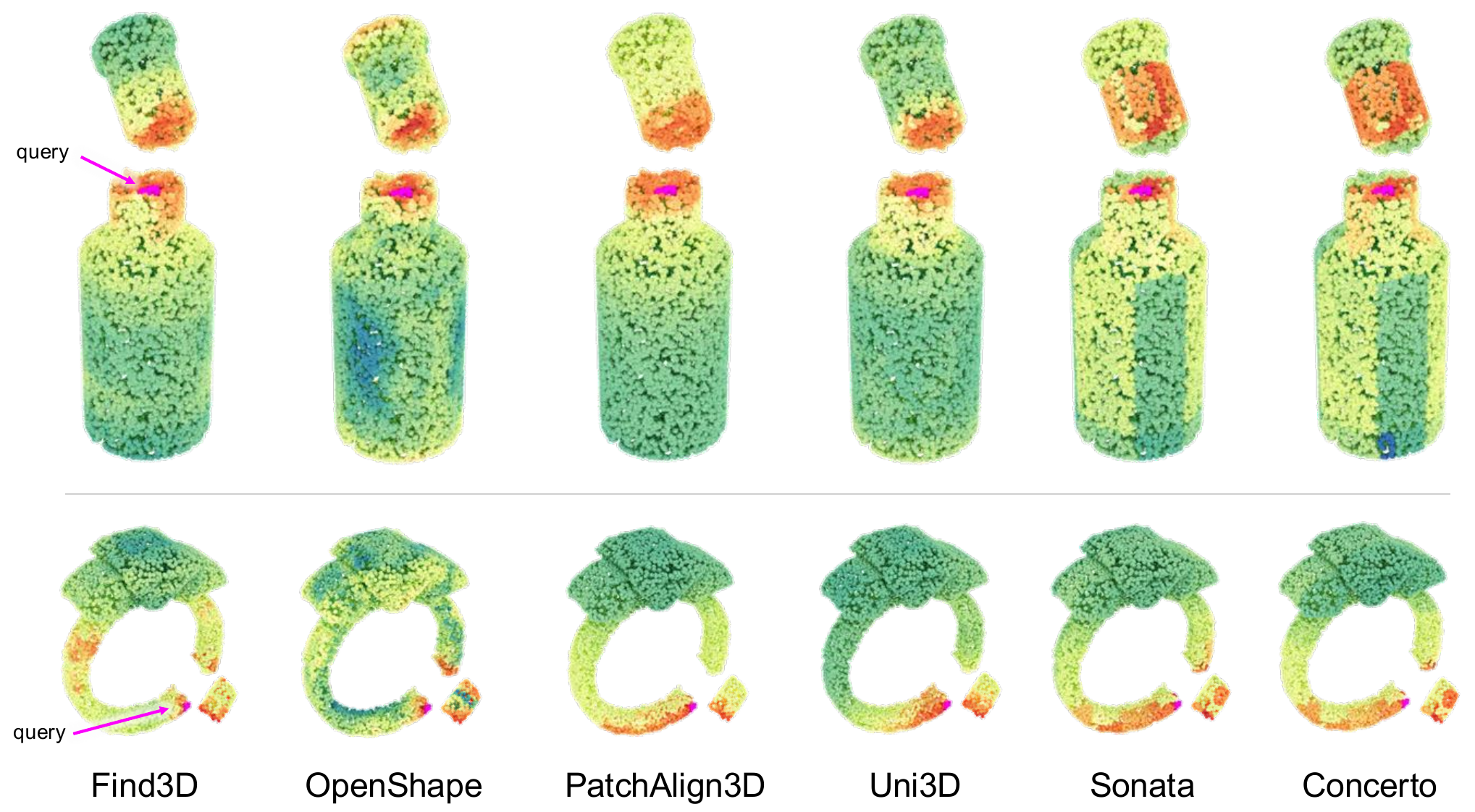}
    \caption{\textbf{Cross-part feature similarity visualization.} For each teacher, we select a query point (pink) on the mating surface of one fragment and color all remaining points by cosine similarity in the frozen feature space (red: high, green: low). Top: bottle, bottom: ring.}
    \vspace{-10pt}
    \label{fig:similarity}
\end{figure}

\subsection{Feature Similarity Visualization}
To provide a more intuitive view of what each teacher representation encodes, Fig.~\ref{fig:similarity} visualizes cross-part feature similarity for representative two-part objects from Breaking Bad-Everyday. Given a query point (pink) on the mating surface of one fragment, we color all remaining points by their cosine similarity to the query in the teacher's frozen feature space. Teachers vary considerably in how well they localize the corresponding mating region on the opposing fragment. Uni3D-L produces the most sharply concentrated responses, while others highlight the contact area to varying degrees.

\begin{figure}[t]
    \centering
    \includegraphics[width=\linewidth]{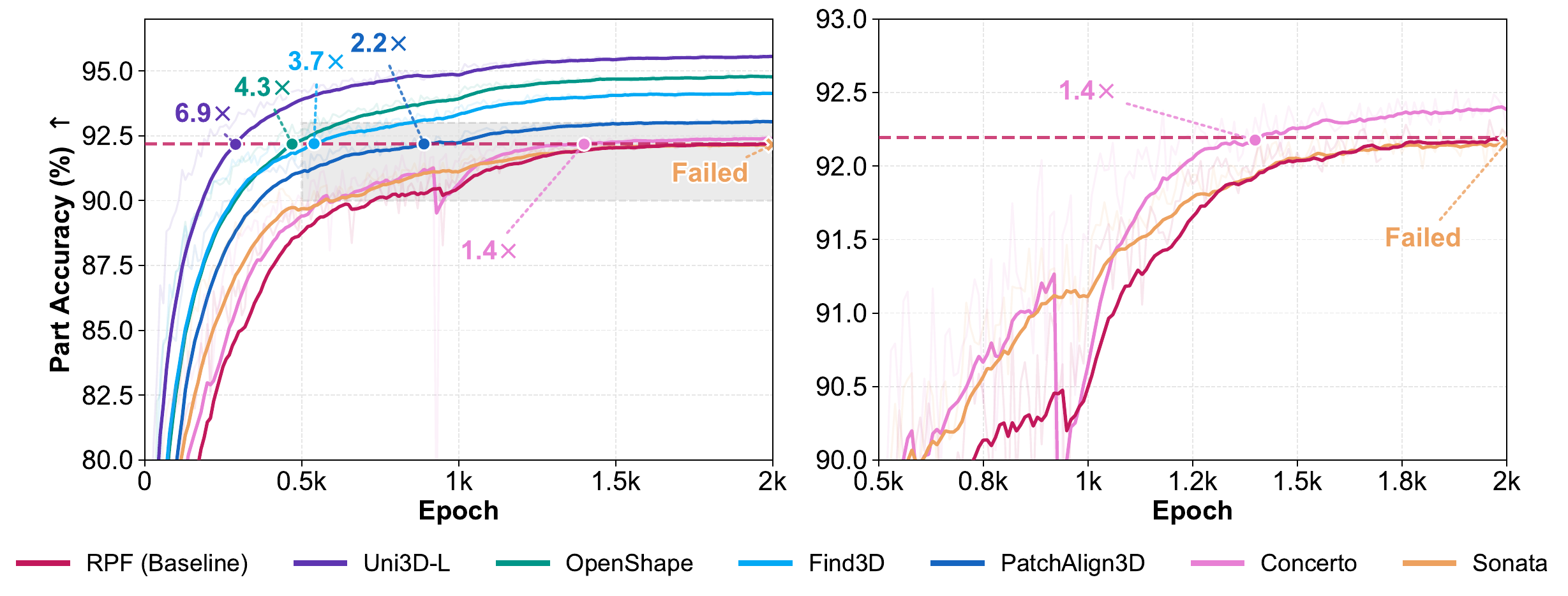}
    \caption{\textbf{Convergence comparison across teachers on Breaking Bad.} We report validation Part Accuracy over training epochs when distilling from each teacher using the CKA objective. Uni3D-L achieves the fastest convergence and highest final accuracy. The right panel provides a zoomed view of the shaded region.}
    \label{fig:convergence_scene_centric}
\end{figure}
\vspace{-2mm}

\subsection{Downstream Convergence}
Figure~\ref{fig:convergence_scene_centric} compares validation Part Accuracy curves on Breaking Bad-Everyday when distilling from each teacher using the CKA objective (Eq.~\mainsec{10}). Uni3D-L achieves the fastest convergence and the highest final accuracy, reaching the baseline's peak performance approximately $6.9\times$ earlier. OpenShape, Find3D, and PatchAlign3D also surpass the RPF baseline, reaching its peak performance $2.2\times$ to $4.3\times$ faster.  Interestingly, Sonata and Concerto do not provide meaningful improvements within the full training schedule, with Concerto yielding only marginal gains and Sonata struggling to reach the baseline's peak accuracy. This aligns with the analyses from above. The right panel of Fig.~\ref{fig:convergence_scene_centric} provides a zoomed view of this region.

Overall, these analyses provide converging support for our choice of Uni3D as the default teacher, which consistently ranks among the top across all three analyses. The scene-centric encoders do not yield reliable improvements in any of the analyses, possibly due to the domain gap between scene-level pretraining and the object-level geometry that assembly relies on.

\subsection{Statistical Validation of Correlation Analysis}
\label{sec:supp_correlation_stats}

\setlength{\columnsep}{4mm}
\begin{wraptable}{r}{0.45\linewidth}
\centering
\vspace{-8mm}
\resizebox{\linewidth}{!}{%
\begin{tabular}{lccc}
\toprule
\textbf{Probe} & $r$ & \textbf{95\% CI} & $r_\text{partial}$ \\
\midrule
\textbf{Seg F1} & $\mathbf{+0.94}$ & $\mathbf{[+0.52,\,+0.99]}$ & $\mathbf{+0.86}$ \\
Part Silh.      & $+0.81$          & $[-0.01,\,+0.98]$          & $+0.59$ \\
LDS             & $+0.51$          & $[-0.52,\,+0.93]$          & $-0.14$ \\
Cls Acc         & $-0.04$          & $[-0.83,\,+0.80]$          & $-0.59$ \\
\bottomrule
\end{tabular}%
}
\vspace{-3mm}
\caption{Fisher 95\% CIs and partial $r$ controlling for \#params.}
\label{tab:supp_teacher_ci}
\vspace{-6mm}
\end{wraptable}
In Sec.~\mainsec{4.1} of the main paper, we report Pearson correlations between teacher probe scores and downstream Part Accuracy over six teacher models. Given the small sample size ($n{=}6$), we provide Fisher-transformed 95\% confidence intervals and partial correlations controlling for teacher parameter count (Tab.~\ref{tab:supp_teacher_ci}).

Seg F1 is statistically supported ($p{=}0.006$, exact); LDS has a wide CI as expected from the small $n$, and we tone down its language in the main paper accordingly while preserving the ordering Seg F1 $\succ$ Part Silh $\succ$ LDS $\succ$ Cls Acc. Partial correlation controlling for \#params leaves Seg F1 essentially unchanged; Cls Acc flips negative, suggesting that geometry/contact probes outperform semantic ones independently of teacher scale.

\subsection{Isolating CKA's Contribution Across Teacher Strengths}
\label{sec:supp_cka_isolation}

\setlength{\columnsep}{4mm}
\begin{wraptable}[9]{r}{0.35\linewidth}
\centering
\vspace{-8mm}
\resizebox{\linewidth}{!}{%
\begin{tabular}{lccc}
\toprule
\textbf{Teacher} & $\mathcal{L}_{\text{cos-dist}}$ & $\mathcal{L}_{\text{CKA}}$ & $\Delta$ \\
\midrule
\multicolumn{4}{l}{\textit{\textbf{Strong teachers}}} \\
Uni3D-G      & 72.2 & \textbf{73.3} & \cellcolor{green!25}$+1.1$ \\
Uni3D-L      & 68.9 & \textbf{71.5} & \cellcolor{green!56}$\mathbf{+2.6}$ \\
\midrule
\multicolumn{4}{l}{\textit{\textbf{Weak teachers}}} \\
Find3D       & 67.7 & \textbf{68.4} & \cellcolor{green!15}$+0.7$ \\
PatchAlign3D & 66.0 & \textbf{68.4} & \cellcolor{green!55}$\underline{+2.4}$ \\
\bottomrule
\end{tabular}%
}
\vspace{-3mm}
\caption{CKA vs.\ Cos-dist (PA~$\uparrow$) on TwoByTwo across teachers of varying strength.}
\label{tab:supp_cka_vs_cos}
\vspace{-8mm}
\end{wraptable}
To test whether CKA's benefit derives from its relational form or from better exploitation of strong teacher features, we compare the top-2 (strong) and bottom-2 (weak) teachers ranked by PA on TwoByTwo (Tab.~\ref{tab:supp_cka_vs_cos}).
CKA improves over Cos-dist on all teachers, including the weak ones. If the gain came from better use of a strong teacher, we would expect it to shrink as the teacher gets weaker. This is not what we see; the \textit{relational form} contributes regardless of teacher capacity.
\section{Additional Implementation Details for TORA}
\label{sec:supp_implementation}
\medbreak\noindent\textbf{Point-wise Features from 3D Point Encoders.}
Our token-wise alignment and linear probing tasks require dense point-level features~\cite{liu2023openshape,yue2025litept,zhou2023uni3d,ma2025find}, whereas several teacher models output only coarse patch-level tokens or subsampled point features. To obtain full-resolution features on the input point cloud of size $N$, we propagate the teacher outputs using inverse distance weighting (IDW) interpolation, following the scheme commonly used in PointNet++~\cite{qi2017pointnet++}. Specifically, for each of the $N$ input points, we identify its $k{=}3$ nearest patch centers or subsampled points in Euclidean space, and compute its dense feature as the distance-weighted average of their features, with weights inversely proportional to spatial distance. We apply this procedure to all teachers that do not natively produce dense point-wise features, including Uni3D, OpenShape, PatchAlign3D, and Find3D. This simple and effective propagation scheme ensures that every point in the student flow is aligned with a corresponding structural feature from the frozen teacher.

\medbreak\noindent\textbf{Alignment Head.}
Following the REPA~\cite{yu2024representation} implementation, we parameterize the projector $\phi$ in Eq.~\mainsec{4}, which maps student features into the teacher feature space, as a 3-layer MLP. The first two layers each consist of a linear projection followed by a SiLU activation, while the final layer is a linear projection without activation. The input and hidden dimensions are set to the student's intermediate feature dimension $D=1\text{,}536$, and the output dimension is set to the teacher feature dimension $D_f$. This lightweight head is used only during training and discarded afterward, thereby incurring zero inference overhead.

\medbreak\noindent\textbf{Discussion on the CKA Subsampling Size $n$.}
As described in Sec.~\mainsec{3.3}, we randomly subsample $n$ token indices per batch to keep the $n{\times}n$ Gram matrices tractable. We study this design choice by training with $n \in \{256, 512, 1024, 2048\}$
\begin{wrapfigure}{r}{0.4\textwidth}
    \centering
    \vspace{-6mm}
    \includegraphics[width=\linewidth]{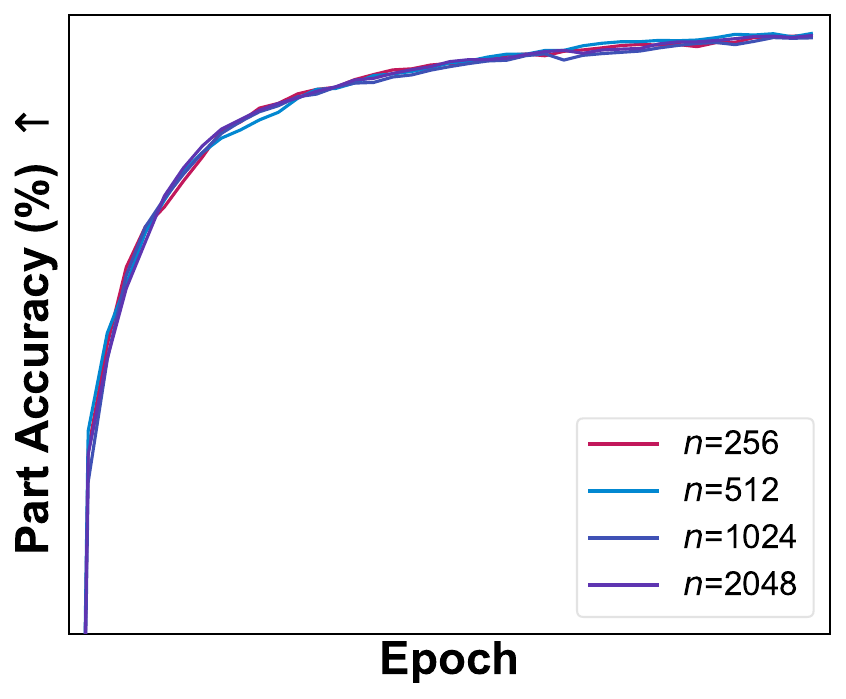}
    \vspace{-6mm}
    \caption{\textbf{Ablation on CKA subsampling size $n$.} Validation Part Accuracy curves on Breaking Bad for $n \in \{256, 512, 1024, 2048\}$. All settings converge nearly identically, confirming robustness to the choice of $n$.}
    \label{fig:cka_n}
\vspace{-8mm}
\end{wrapfigure}
and tracking validation Part Accuracy on the Breaking Bad dataset. 
As shown in Fig.~\ref{fig:cka_n}, all four settings exhibit nearly identical convergence behavior, indicating that the CKA objective is robust to the subsampling size across this range. 
This robustness likely arises because even a moderate number of randomly sampled tokens provides a sufficiently faithful estimate of the full Gram-matrix structure. Based on this observation, we adopt $n{=}1\text{,}024$ as a conservative default.

Although smaller values such as $n{=}256$ are marginally cheaper, the practical difference is negligible. As shown in Tab.~\mainsec{4}, CKA alignment with offline-cached teacher features adds only ${\sim}$3\,ms per step ($+2.8\%$), largely independent of $n$, because the dominant cost comes from teacher-feature lookup and the projector forward pass rather than from the Gram-matrix computation itself. We therefore favor a moderately large $n$ to ensure a stable similarity estimate throughout the full 2{,}000-epoch training schedule.

\section{Evaluation Protocol}
\label{sec:supp_eval}

In the main paper, we report all results under a \textit{unified evaluation protocol} applied consistently across all methods. Here, we describe this protocol in detail and clarify our motivation for re-evaluating all baselines rather than directly adopting numbers from prior work.

\medbreak\noindent\textbf{Motivation for unified re-evaluation.}
During our experimental setup, we found that the evaluation protocol reported in the main RPF paper~\cite{sun2025rectified} could not be reproduced from the official codebase and released checkpoints. In particular, the main paper reports anchor-fixed results, whereas the official implementation runs in an anchor-free setting and evaluates rotation and translation errors with ICP-based residual pose estimation~\cite{besl1992icp}, a detail not documented in the paper. With the released code and pretrained checkpoints, we could reproduce the anchor-free results reported in the RPF supplementary material, but not the anchor-fixed numbers in the main paper. To ensure fairness and reproducibility, we therefore have re-evaluated all methods under a single protocol in the main paper, using official pretrained checkpoints whenever available. Specifically, we adopted the ICP-based evaluation procedure provided in the RPF codebase and applied it uniformly to all methods under the anchor-fixed setting. The anchor-free RPF results in Sec.~\ref{sec:supp_anchor_free} are quoted directly from the original paper. Under both evaluation protocols, our method consistently outperforms RPF.

\medbreak\noindent\textbf{Notation.} 
Throughout this section, let $\mathbf{P}_k \in \mathbb{R}^{N_k \times 3}$ denote the input point cloud of part $k$, $\mathbf{P}^*_k$ its ground-truth assembled placement, and $\hat{\mathbf{T}}_k = (\hat{R}_k, \hat{t}_k) \in \text{SE}(3)$ the predicted pose, yielding the predicted placement $\hat{\mathbf{P}}_k = \mathbf{P}_k \hat{R}_k^\top + \hat{t}_k$.

\medbreak\noindent\textbf{Part Accuracy.}
Part Accuracy (PA) directly evaluates the quality of the predicted placement without any post-hoc alignment. PA measures the fraction of parts whose bidirectional Chamfer distance between $\hat{\mathbf{P}}_k$ and $\mathbf{P}^*_k$ falls strictly below a threshold $\tau$:
\begin{equation}
    \mathrm{PA} = \frac{1}{K}\sum_{k=1}^{K} \mathbbm{1}\!\left[\,\mathrm{CD}(\hat{\mathbf{P}}_k,\, \mathbf{P}^*_k) < \tau\,\right],
\end{equation}
where $\mathrm{CD}(\cdot, \cdot)$ denotes the Chamfer distance and $\tau = 0.01$.

\medbreak\noindent\textbf{Rotation and Translation Error.}
Unlike PA, which is computed on point clouds and therefore inherently invariant to pose symmetries, rotation and translation errors require comparing pose parameters directly. A common approach is to compare predicted poses against ground-truth annotations. While effective in many settings, this can be problematic for \textit{geometrically symmetric} parts: a cylinder admits infinitely many valid rotations about its axis, all producing identical placements, yet direct comparison penalises all but the single annotated rotation. This issue is well recognized in the 6D object pose estimation community, where symmetry-aware metrics such as ADD-S~\cite{xiang2017posecnn} and MSSD~\cite{hodan2018bop} have been proposed to evaluate placement quality via surface distances rather than direct pose comparison.

We follow the same principle in the shape assembly setting. Rather than comparing poses directly against ground-truth annotations, we estimate the residual pose error via ICP~\cite{besl1992icp}. The key insight is that ICP converges to the nearest local minimum: for a correctly placed symmetric part, the residual transform will be $(R_\epsilon, t_\epsilon) \approx (\mathbf{I}, \mathbf{0})$ regardless of which symmetry-equivalent rotation was predicted, making the resulting RE and TE invariant to symmetric ambiguity.

Concretely, for each non-anchor part $k$, we run point-to-point ICP from $\mathbf{P}^*_k$ to $\hat{\mathbf{P}}_k$, recovering the residual rigid transform $(R_\epsilon, t_\epsilon)$. We then measure rotation errors using the geodesic distance on $\mathrm{SO}(3)$,
\begin{equation}
    \mathrm{RE}_k = \cos^{-1}\!\left(\frac{\mathrm{tr}(R_\epsilon) - 1}{2}\right),
\end{equation}
which is coordinate-free, singularity-free, and corresponds to the Riemannian metric on $\mathrm{SO}(3)$~\cite{sun2025rectified}. Translation errors are computed as the root-mean-square of the three residual translation components, rescaled to metric units:
\begin{equation}
    \mathrm{TE}_k = \sqrt{\frac{t_{\epsilon,x}^2 + t_{\epsilon,y}^2 + t_{\epsilon,z}^2}{3}} \;\cdot\; s,
\end{equation}
where $s$ is the per-sample normalization scale factor. Errors expressed in normalized coordinates are not comparable across objects of different sizes, whereas metric-scale TE reflects the true physical misalignment regardless of object extent. Both RE and TE are averaged over all $K$ valid parts per sample.

\newcolumntype{C}{>{\centering\arraybackslash}p{1.8cm}} 

\begin{table}[!t]
\centering
\caption{\textbf{Anchor-free evaluation on shape assembly benchmarks.} We compare against RPF~\cite{sun2025rectified} under both anchor-fixed and anchor-free protocols. The \colorbox{blue!20}{\textbf{best}} results are highlighted per column.}
\resizebox{\linewidth}{!}{
\begin{tabular}{l l CCC CCC CC}
\toprule
& & \multicolumn{3}{c}{\textbf{Breaking Bad}~\cite{sellan2022breaking}} 
& \multicolumn{3}{c}{\textbf{PartNet-Assembly}~\cite{xu2025spaformer}} 
& \multicolumn{2}{c}{\textbf{TwoByTwo}~\cite{qi2025two}} \\
\cmidrule(lr){3-5} \cmidrule(lr){6-8} \cmidrule(lr){9-10}
\textbf{Protocol} & \textbf{Method}
& PA (\%) $\uparrow$ & RE ($^\circ$) $\downarrow$ & TE (cm) $\downarrow$ 
& PA (\%) $\uparrow$ & RE ($^\circ$) $\downarrow$ & TE (cm) $\downarrow$ & RE ($^\circ$) $\downarrow$ & TE (cm) $\downarrow$ \\
\midrule
\multirow{2}{*}{Anchor-fixed}
& RPF~\cite{sun2025rectified} & 93.2 & 16.0 & 4.3 & 59.8 & 46.2 & 21.5  & 15.8 & 11.9 \\
& \textbf{Ours}$_\textrm{CKA}$ & \best{95.7} & \best{8.6} & \best{2.1} & \best{69.1} & \best{40.8} & \best{18.8} & \best{10.0} & \best{7.6} \\
\midrule

\multirow{2}{*}{Anchor-free}
& RPF~\cite{sun2025rectified}            & 90.2 & 17.4 & 8.0  & 45.3 & 47.3 & 40.5  & 15.2 & 24.2 \\
& \textbf{Ours}$_\textrm{CKA}$          & \best{94.0} & \best{11.5}  & \best{5.1}  & \best{52.1} & \best{44.5} & \best{38.6} &  \best{14.9} & \best{21.4} \\

\bottomrule
\end{tabular}
}
\label{tab:anchor_free}
\end{table}

\subsection{Anchor-free Evaluation}
\label{sec:supp_anchor_free}
In the standard anchor-fixed protocol, one part is assigned a known pose ($\hat{\mathbf{T}}_1 = \mathbf{I}$) at test time, providing the model with a free global reference frame. As noted by Sun~\etal~\cite{sun2025rectified}, this introduces a positive bias: the anchor eliminates global positional and rotational drift for all connected parts, and the reported numbers become anchor-dependent.

Following RPF~\cite{sun2025rectified}, we therefore also evaluate under an \emph{anchor-free} protocol, where the model receives no privileged pose information for any part; the anchor's point cloud is centered to its own center of mass and randomly rotated, exactly as for every other part. Table~\ref{tab:anchor_free} compares both protocols, with RPF anchor-free numbers copied directly from the original paper. Our method shows consistent improvements across all metrics under both protocols, demonstrating robust assembly quality even without privileged anchor information.
\section{Experimental Validation}
\label{sec:supp_validation}

We provide additional analyses that validate the statistical significance of our main results, and investigate the NT-Xent failure mode on inter-object assembly.

\subsection{Variance and Statistical Significance}
\label{sec:supp_variance}

To verify that the gains reported in Tab.~\mainsec{1} of the main paper are not attributable to seed variance, we re-run all entries with $N{=}3$ random seeds and report mean $\pm$ standard deviation alongside paired $t$-tests comparing CKA against Cos-dist on shared seeds (Tab.~\ref{tab:supp_variance}).
On PartNet-Assembly and TwoByTwo, where headroom above the baseline exists, CKA's Part Accuracy gains over Cos-dist ($+1.41$, $+2.80$) exceed the larger seed standard deviation by $8.8\times$ and $9.0\times$ respectively, with consistent reductions in rotation and translation errors.
On Breaking Bad-Everyday, both alignment variants exceed 95\% PA, a regime where the metric saturates and limits differentiation; nevertheless, RE and TE still show statistically significant improvements ($p{=}0.005$ and $p{=}0.003$).
Across all benchmarks, every CKA-vs-Cos-dist comparison reaches $p{<}0.05$, confirming that the reported gains are real and reproducible.

\begin{table}[h]
\centering
\caption{\textbf{Variance and significance analysis.} Mean $\pm$ std over 3 seeds, with paired $t$-test $p$-values for CKA vs.\ Cos-dist.}
\label{tab:supp_variance}
\vspace{-2mm}
\resizebox{0.75\linewidth}{!}{%
\begin{tabular}{lcrrrr}
\toprule
\textbf{Dataset} & \textbf{Metric} & $\mathcal{L}_{\text{cos-dist}}$ & $\mathcal{L}_{\text{CKA}}$ & $\Delta$ & $p$ \\
\midrule
\multirow{3}{*}{Breaking Bad - Everyday}
  & PA $\uparrow$   & $95.70_{\pm 0.03}$ & $95.71_{\pm 0.01}$ & $+0.01$ & $0.64$ \\
  & RE $\downarrow$ & $9.01_{\pm 0.03}$  & $8.64_{\pm 0.07}$  & $-0.37$ & $\mathbf{0.005}$ \\
  & TE $\downarrow$ & $2.20_{\pm 0.02}$  & $2.08_{\pm 0.01}$  & $-0.12$ & $\mathbf{0.003}$ \\
\midrule
\multirow{3}{*}{PartNet-Assembly}
  & PA $\uparrow$   & $67.70_{\pm 0.16}$ & $69.11_{\pm 0.04}$ & $+1.41$ & $\mathbf{0.006}$ \\
  & RE $\downarrow$ & $41.58_{\pm 0.08}$ & $40.82_{\pm 0.05}$ & $-0.76$ & $\mathbf{0.002}$ \\
  & TE $\downarrow$ & $19.15_{\pm 0.05}$ & $18.80_{\pm 0.01}$ & $-0.35$ & $\mathbf{0.008}$ \\
\midrule
\multirow{3}{*}{TwoByTwo}
  & PA $\uparrow$   & $68.76_{\pm 0.31}$ & $71.56_{\pm 0.08}$ & $+2.80$ & $\mathbf{0.003}$ \\
  & RE $\downarrow$ & $11.84_{\pm 0.18}$ & $10.04_{\pm 0.40}$ & $-1.80$ & $\mathbf{0.011}$ \\
  & TE $\downarrow$ & $9.39_{\pm 0.11}$  & $7.67_{\pm 0.44}$  & $-1.72$ & $\mathbf{0.027}$ \\
\bottomrule
\end{tabular}%
}
\end{table}

\vspace{-3mm}
\subsection{NT-Xent Failure Mode Analysis}
\label{sec:supp_ntxent}

\setlength{\columnsep}{4mm}
\begin{wraptable}{r}{0.4\linewidth}
\centering
\vspace{-7mm}
\resizebox{\linewidth}{!}{%
\begin{tabular}{lccc}
\toprule
\textbf{Benchmark} & \makecell{Category\\shift} & \makecell{Inter-\\object} & $\Delta_\text{PA}$ \\
\midrule
BBad-Everyday    & \ding{55} & \ding{55} & \cellcolor{red!10}$-0.3$ \\
PartNet-Assm.    & \ding{55} & \ding{55} & \cellcolor{green!50}$+5.6$ \\
\midrule
BBad-Artifact    & \ding{51} & \ding{55} & \cellcolor{red!20}$-1.3$ \\
FRACTURA         & \ding{51} & \ding{55} & \cellcolor{green!35}$+3.6$ \\
Fantastic Breaks & \ding{51} & \ding{55} & \cellcolor{green!15}$+0.7$ \\
\midrule
TwoByTwo         & \ding{55} & \ding{51} & \cellcolor{red!50}$\mathbf{-5.0}$ \\
\bottomrule
\end{tabular}%
}
\vspace{-3mm}
\caption{NT-Xent failure analysis. $\Delta_\text{PA}$ is relative to RPF.}
\label{tab:supp_ntxent}
\vspace{-7mm}
\end{wraptable}
In Sec.~\mainsec{5}, we observe that NT-Xent degrades performance on TwoByTwo. To determine whether this is driven by category shift or the inter-object setting, we compare NT-Xent's $\Delta_\text{PA}$ relative to RPF across six benchmarks (Tab.~\ref{tab:supp_ntxent}).
NT-Xent fluctuates around the RPF baseline on all five intra-object benchmarks, irrespective of whether category shift is present. However, it degrades sharply on TwoByTwo ($-5.0$ PA), the only inter-object benchmark. 
This indicates that the failure is driven by the \textbf{inter-object factor}: NT-Xent's negatives include tokens from the \emph{other} object, which the teacher (trained on isolated objects) already separates. Enforcing further discriminability between these tokens suppresses the cross-object mating cues that are essential for inter-object assembly.

\section{Failure Cases and Future Directions}
\label{sec:suppl_failure}

\begin{figure}[h]
    \centering
    \includegraphics[width=0.98\textwidth]{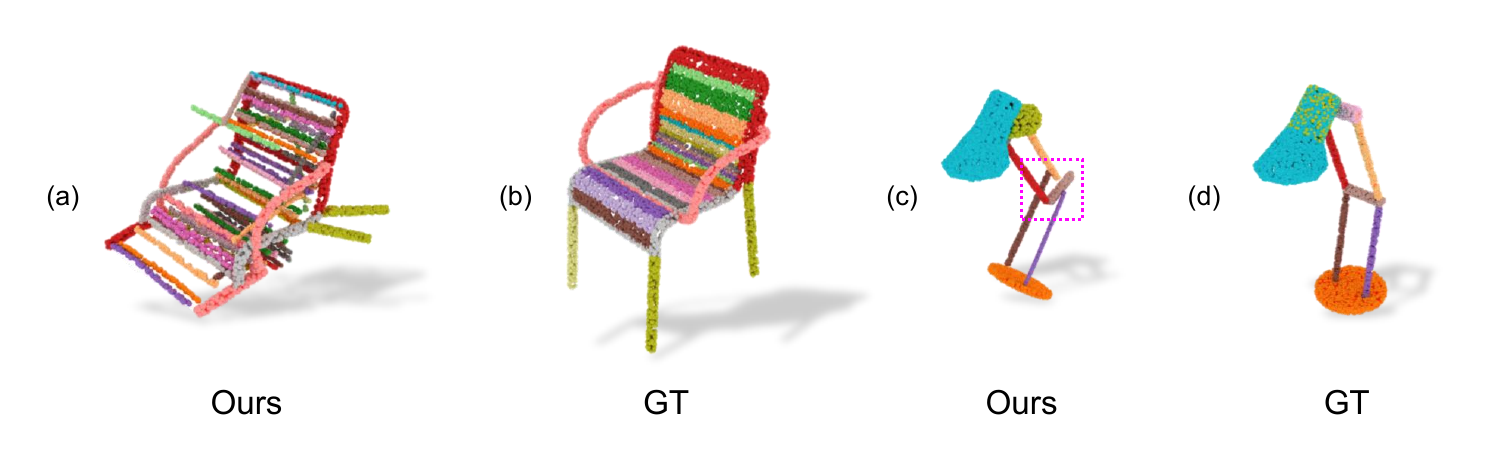}
    \caption{\textbf{Part repetitions and hallucinated discontinuity.} Two failure modes from PartNet are depicted. Left: (a), (b) display the failed assembly of an object with high ambiguity; many symmetric and repeating horizontal bars with minuscule mating surfaces. Right: (c), (d) display plausible pose estimates, but a slight misorientation leads to discontinuity and a non-functional object.}
    \label{fig:suppl_qual_failures}
\end{figure}

Like all current assembly methods, TORA operates most reliably when parts are geometrically distinctive. In assemblies with high part symmetry or repetition such as semantic assembly on PartNet, the inter-part signal becomes underspecified and predicted configurations may possess degraded global visual coherence; see Fig.~\ref{fig:suppl_qual_failures}(a). The representation alignment objective does not directly enforce surface contact, meaning boundary gaps can occasionally arise in otherwise well-posed predictions. These geometric hallucinations are visually salient but may not be reflected proportionately in error metrics; see Fig.~\ref{fig:suppl_qual_failures}(c). As with any teacher-guided framework, the geometric capacity of the chosen teacher model sets a ceiling on representational expressiveness, with the sensitivity to teacher choice remaining an open question. TORA supports various directions for future work including scaling point-wise attention to larger assemblies with denser point clouds, enriching alignment with visual and functional semantics, and reducing inference complexity to support real-time robotic registration.

\section{Additional Qualitative Results}
\label{sec:additional_qual}
\begin{figure}[h]
    \centering
    \includegraphics[width=0.98\textwidth]{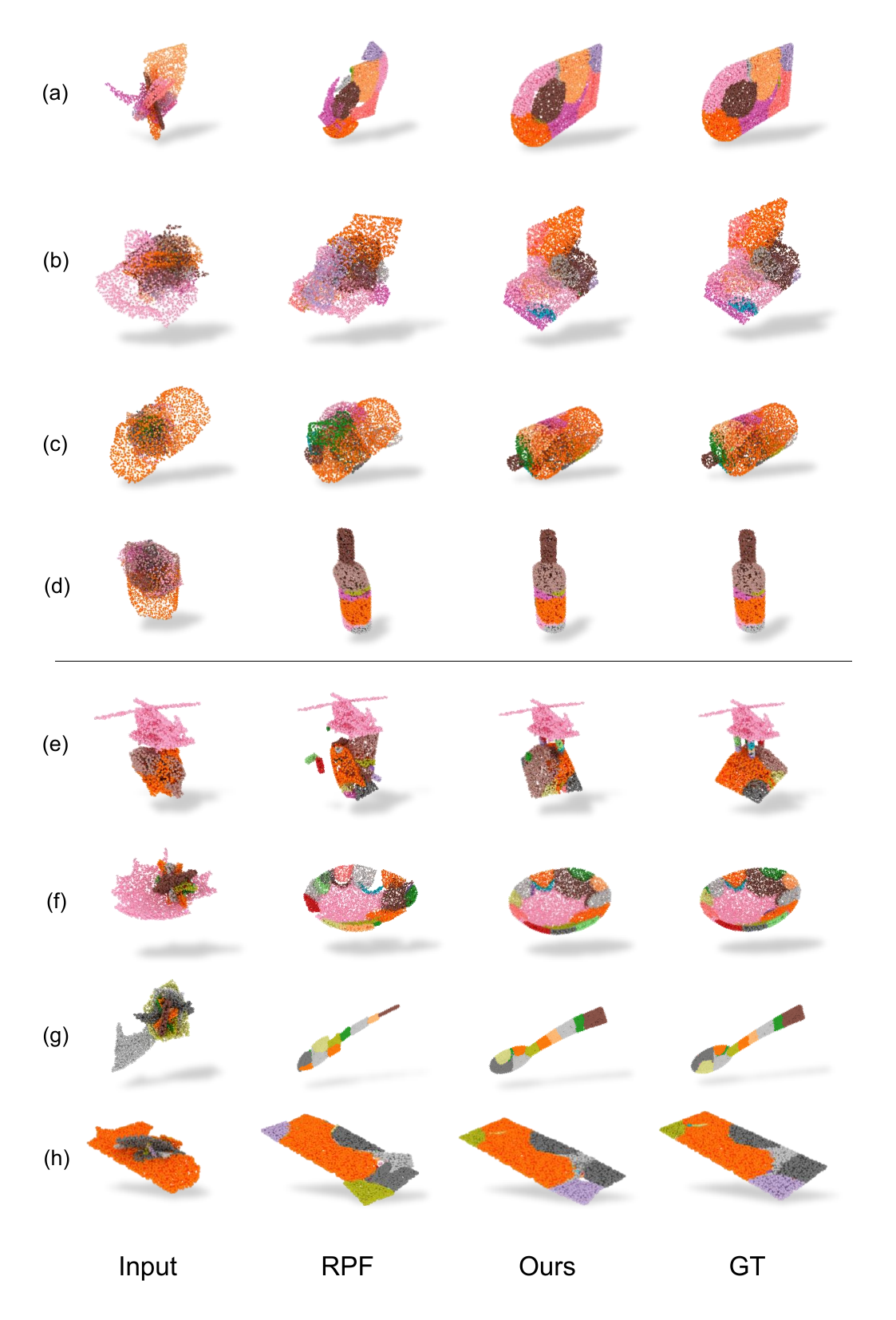}
    \caption{\textbf{Breaking Bad Everyday.} Additional qualitative comparison for \textit{geometric} shape assembly. Top half (a-d) has 2 to 20 parts, bottom half (e-h) has 21 to 33.}
    \label{fig:suppl_qual_everyday}
\end{figure}

\begin{figure}[h]
    \centering
    \includegraphics[width=0.98\textwidth]{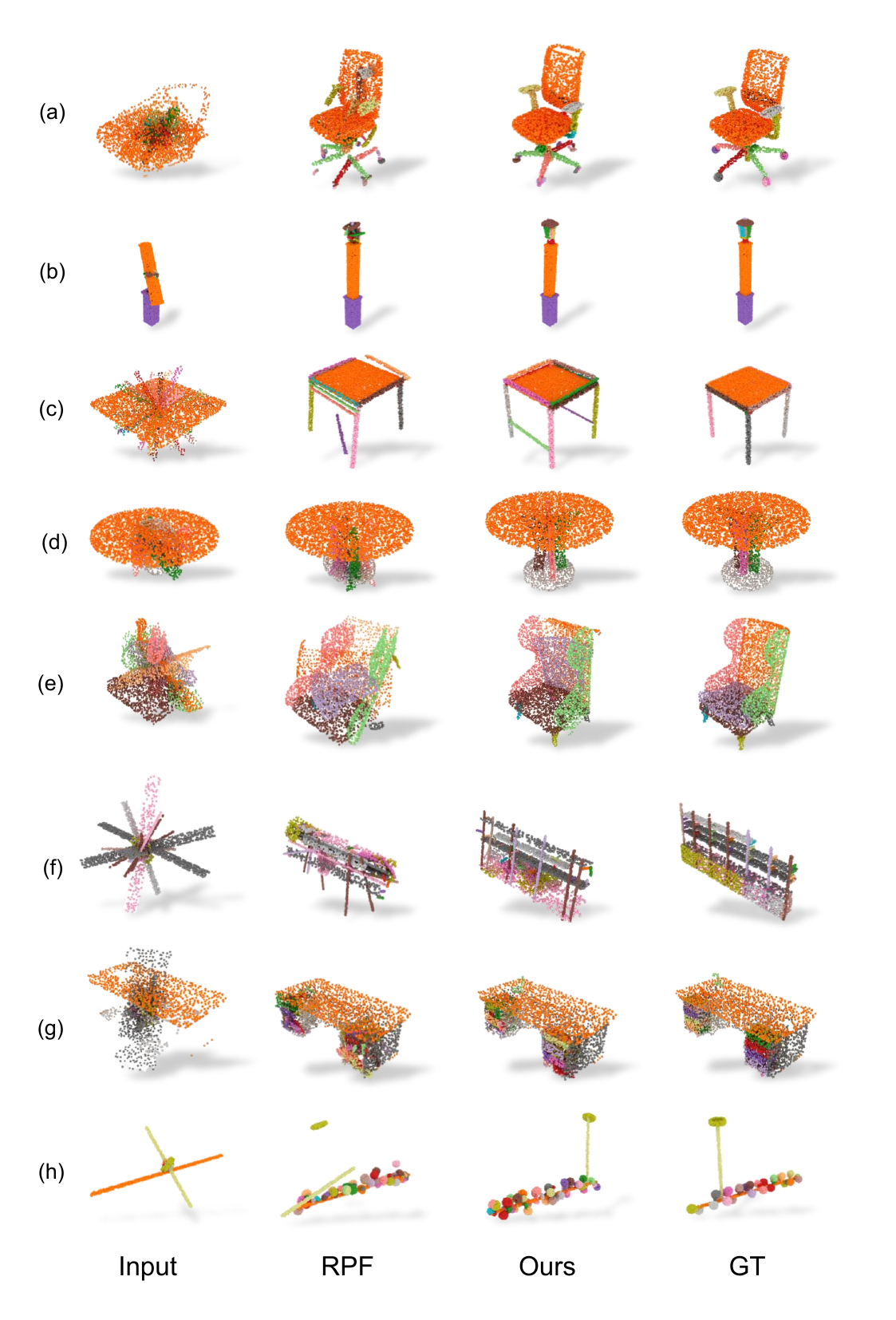}
    \caption{\textbf{PartNet-Assembly.} Additional qualitative comparison for \textit{semantic} shape assembly. Top half (a-d) has 2 to 30 parts, bottom half (e-h) has 31 to 64.}
    \label{fig:suppl_qual_partnet}
\end{figure}

\begin{figure}[h]
    \centering
    \includegraphics[width=0.98\textwidth]{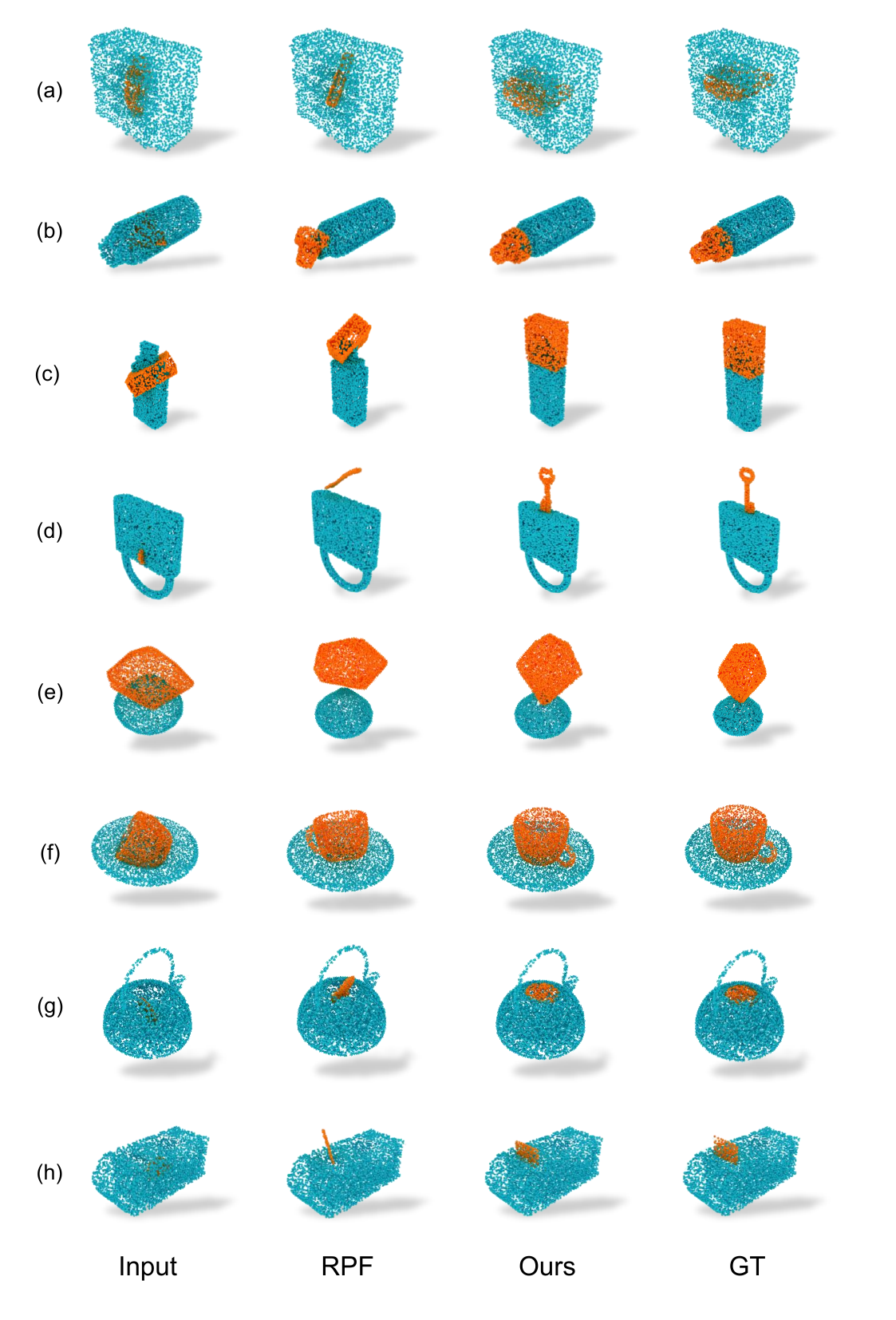}
    \caption{\textbf{TwoByTwo.} Additional qualitative comparison for \textit{inter-object} shape assembly.}
    \label{fig:suppl_qual_twbytwo}
\end{figure}

\begin{figure}[h]
    \centering
    \includegraphics[width=0.98\textwidth]{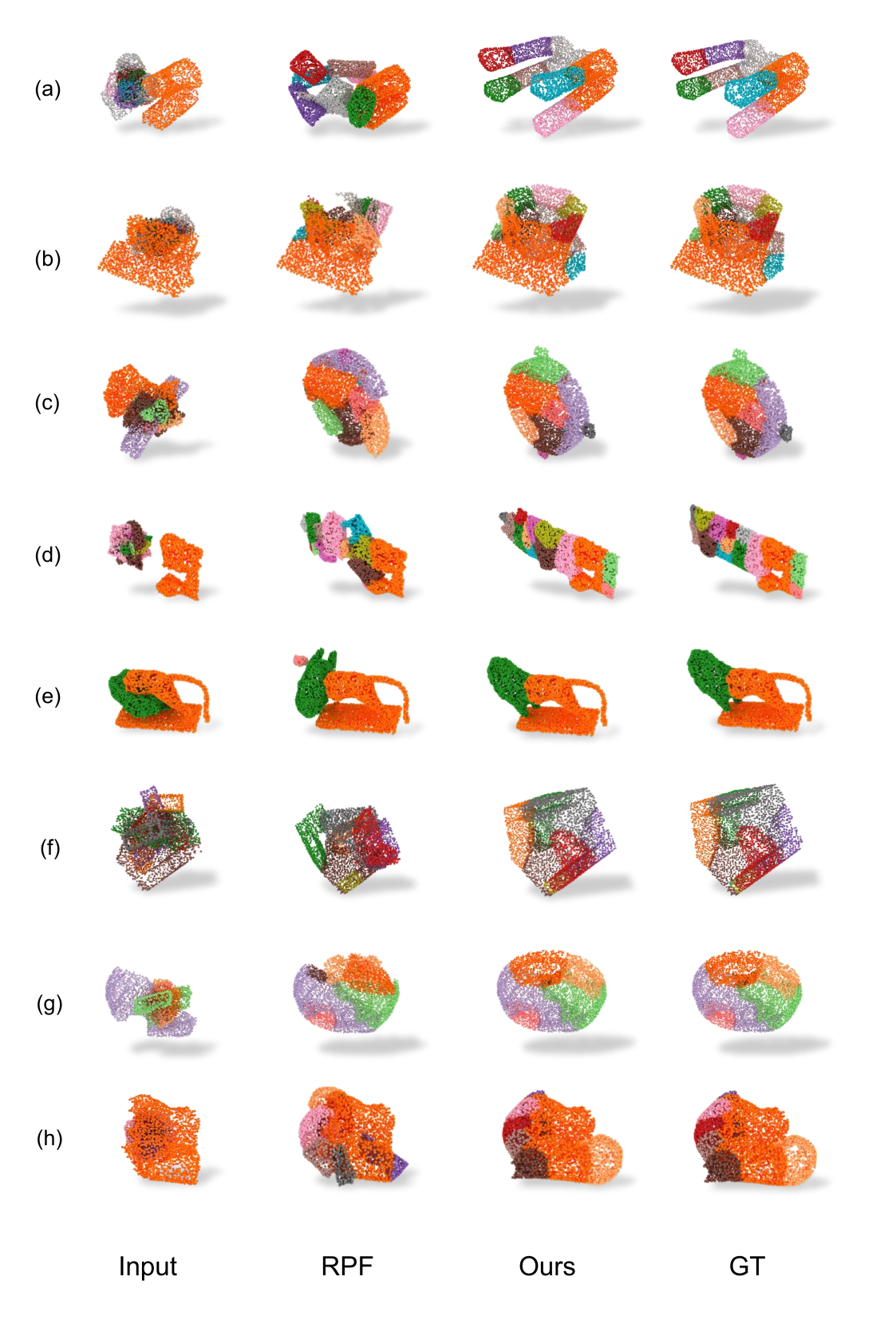}
    \caption{\textbf{Breaking Bad Artifact.} Additional qualitative comparison for zero-shot shape assembly on \textit{synthetic} objects from unseen categories.}
    \label{fig:suppl_qual_artifact}
\end{figure}

\begin{figure}[h]
    \centering
    \includegraphics[width=0.98\textwidth]{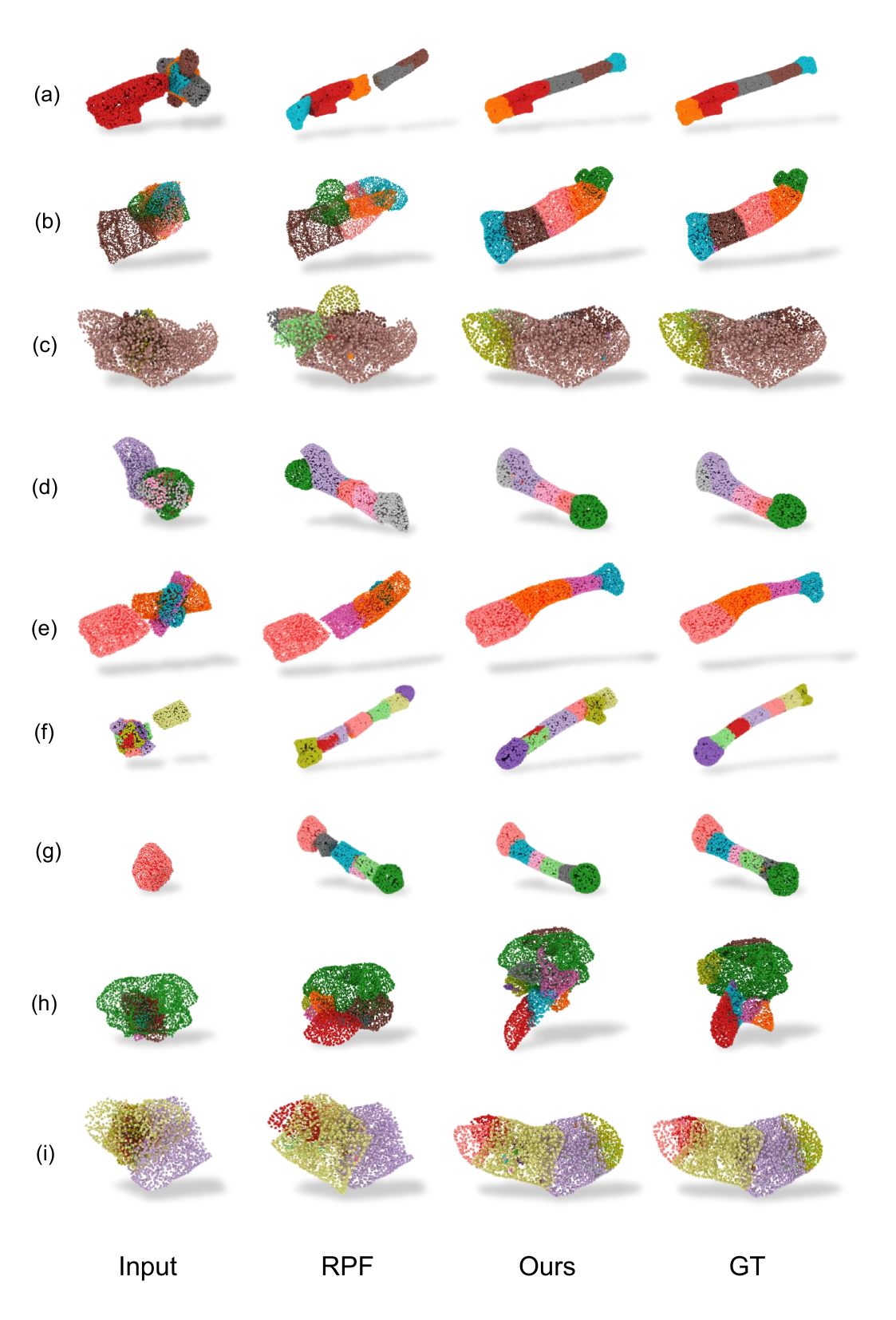}
    \caption{\textbf{FRACTURA.} Additional qualitative comparison for zero-shot shape assembly on objects with \textit{mixed} synthetic and real fractures.}
    \label{fig:suppl_qual_fractura}
\end{figure}

\begin{figure}[h]
    \centering
    \includegraphics[width=0.98\textwidth]{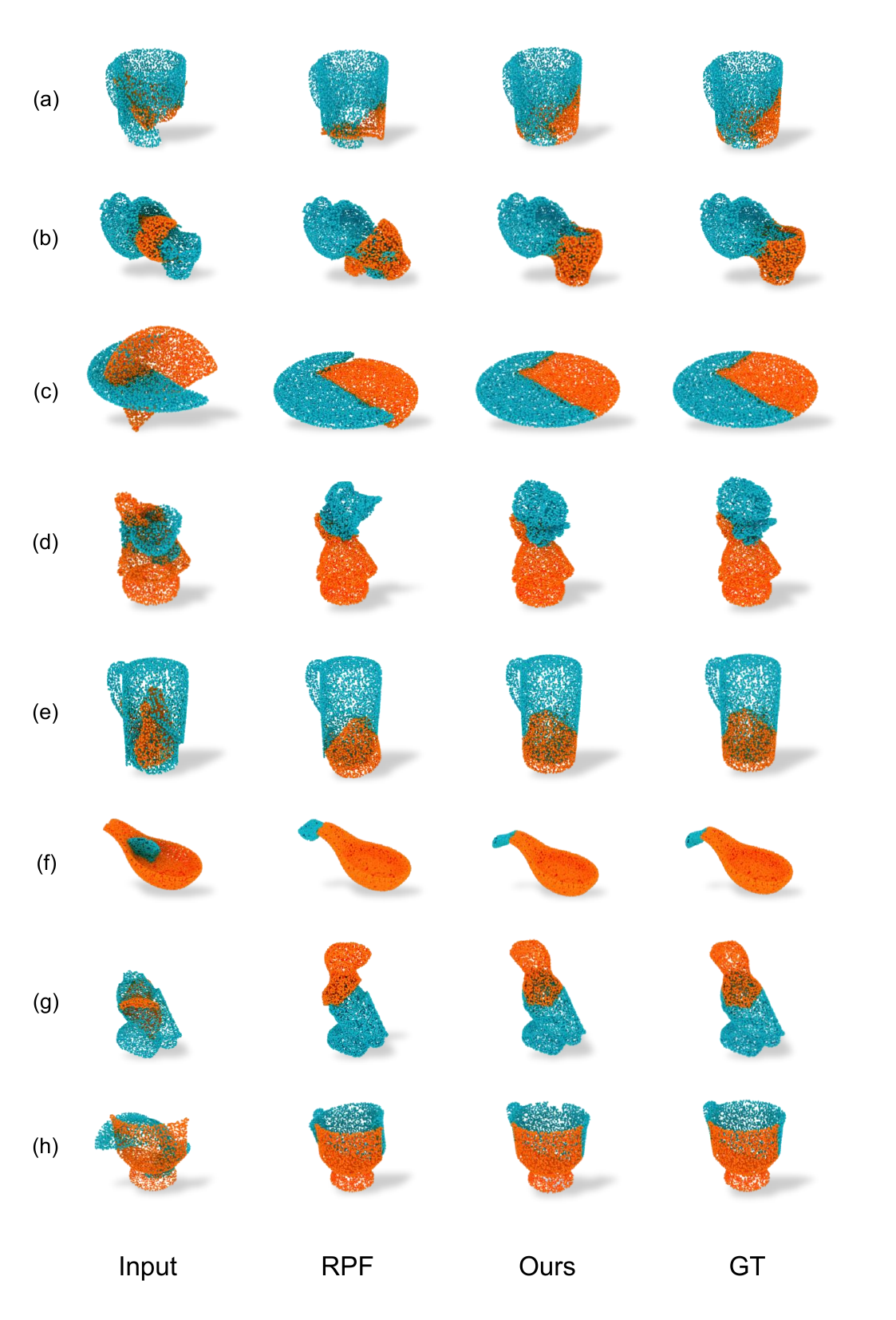}
    \caption{\textbf{Fantastic Breaks.} Additional qualitative comparison for zero-shot shape assembly on \textit{real-world} scanned objects.}
    \label{fig:suppl_qual_fantabreaks}
\end{figure}

We present more qualitative results in Figs.~\ref{fig:suppl_qual_everyday},~\ref{fig:suppl_qual_partnet},~\ref{fig:suppl_qual_twbytwo},~\ref{fig:suppl_qual_artifact},~\ref{fig:suppl_qual_fractura} and~\ref{fig:suppl_qual_fantabreaks}.

%
%
\clearpage
\bibliographystyle{splncs04}
\bibliography{references}

@String(CVPR  = {IEEE Conf. Comput. Vis. Pattern Recog.})

@String(ICCV  = {Int. Conf. Comput. Vis.})

@String(ECCV  = {Eur. Conf. Comput. Vis.})

@String(ICML  = {Int. Conf. Mach. Learn.})

@String(ICLR  = {Int. Conf. Learn. Represent.})

@String(CVPRW = {IEEE Conf. Comput. Vis. Pattern Recog. Worksh.})

@String(TOG   = {ACM Trans. Graph.})

@String(CVPR  = {CVPR})

@String(ICCV  = {ICCV})

@String(ECCV  = {ECCV})

@String(ICML  = {ICML})

@String(ICLR  = {ICLR})

@String(CVPRW = {CVPRW})

@String(TOG   = {ACM TOG})

@inproceedings{lee2025cmnet,
  title={Combinative Matching for Geometric Shape Assembly},
  author={Lee, Nahyuk and Min, Juhong and Lee, Junhong and Park, Chunghyun and Cho, Minsu},
  booktitle={Proceedings of the IEEE/CVF International Conference on Computer Vision},
  pages={9540--9549},
  year={2025}
}

@article{an2025c3g,
  title={C3G: Learning Compact 3D Representations with 2K Gaussians},
  author={An, Honggyu and Jung, Jaewoo and Kim, Mungyeom and Hong, Sunghwan and Kim, Chaehyun and Fukuda, Kazumi and Jeon, Minkyeong and Han, Jisang and Narihira, Takuya and Ko, Hyuna and others},
  journal={arXiv preprint arXiv:2512.04021},
  year={2025}
}

@article{yoon2025visual,
  title={Visual representation alignment for multimodal large language models},
  author={Yoon, Heeji and Jung, Jaewoo and Kim, Junwan and Choi, Hyungyu and Shin, Heeseong and Lim, Sangbeom and An, Honggyu and Kim, Chaehyun and Han, Jisang and Kim, Donghyun and others},
  journal={arXiv preprint arXiv:2509.07979},
  year={2025}
}

@article{lee20253d,
  title={3D Scene Prompting for Scene-Consistent Camera-Controllable Video Generation},
  author={Lee, JoungBin and Jung, Jaewoo and Han, Jisang and Narihira, Takuya and Fukuda, Kazumi and Seo, Junyoung and Hong, Sunghwan and Mitsufuji, Yuki and Kim, Seungryong},
  journal={arXiv preprint arXiv:2510.14945},
  year={2025}
}

@article{yue2025litept,
  title={Litept: Lighter yet stronger point transformer},
  author={Yue, Yuanwen and Robert, Damien and Wang, Jianyuan and Hong, Sunghwan and Wegner, Jan Dirk and Rupprecht, Christian and Schindler, Konrad},
  journal={arXiv preprint arXiv:2512.13689},
  year={2025}
}

@article{han2026geometric,
  title={Geometric Action Model for Robot Policy Learning},
  author={Han, Jisang and Jeon, Seonghu and Jung, Jaewoo and Zurbr{\"u}gg, Ren{\'e} and An, Honggyu and Portela, Tifanny and Hutter, Marco and Pollefeys, Marc and Kim, Seungryong and Hong, Sunghwan},
  journal={arXiv preprint arXiv:2606.17046},
  year={2026}
}

@article{chen2026prose,
  title={PROSE: Training-Free Egocentric Scene Registration with Vision-Language Models},
  author={Chen, Zhiang and Lee, Nahyuk and Sun, Boyang and Kwon, Taein and Pollefeys, Marc and Bauer, Zuria and Hong, Sunghwan},
  journal={arXiv preprint arXiv:2606.16569},
  year={2026}
}

@article{han2025emergent,
  title={Emergent Outlier View Rejection in Visual Geometry Grounded Transformers},
  author={Han, Jisang and Hong, Sunghwan and Jung, Jaewoo and Jang, Wooseok and An, Honggyu and Wang, Qianqian and Kim, Seungryong and Feng, Chen},
  journal={arXiv preprint arXiv:2512.04012},
  year={2025}
}

@inproceedings{hong2024unifying,
  title={Unifying correspondence pose and nerf for generalized pose-free novel view synthesis},
  author={Hong, Sunghwan and Jung, Jaewoo and Shin, Heeseong and Yang, Jiaolong and Kim, Seungryong and Luo, Chong},
  booktitle={Proceedings of the IEEE/CVF Conference on Computer Vision and Pattern Recognition},
  pages={20196--20206},
  year={2024}
}

@article{hong2024pf3plat,
  title={Pf3plat: Pose-free feed-forward 3d gaussian splatting},
  author={Hong, Sunghwan and Jung, Jaewoo and Shin, Heeseong and Han, Jisang and Yang, Jiaolong and Luo, Chong and Kim, Seungryong},
  journal={arXiv preprint arXiv:2410.22128},
  year={2024}
}

@article{cho2021cats,
  title={Cats: Cost aggregation transformers for visual correspondence},
  author={Cho, Seokju and Hong, Sunghwan and Jeon, Sangryul and Lee, Yunsung and Sohn, Kwanghoon and Kim, Seungryong},
  journal={Advances in Neural Information Processing Systems},
  volume={34},
  pages={9011--9023},
  year={2021}
}

@article{hong2022neural,
  title={Neural matching fields: Implicit representation of matching fields for visual correspondence},
  author={Hong, Sunghwan and Nam, Jisu and Cho, Seokju and Hong, Susung and Jeon, Sangryul and Min, Dongbo and Kim, Seungryong},
  journal={Advances in Neural Information Processing Systems},
  volume={35},
  pages={13512--13526},
  year={2022}
}

@article{hong2024unifying2,
  title={Unifying feature and cost aggregation with transformers for semantic and visual correspondence},
  author={Hong, Sunghwan and Cho, Seokju and Kim, Seungryong and Lin, Stephen},
  journal={arXiv preprint arXiv:2403.11120},
  year={2024}
}

@article{cho2022cats++,
  title={Cats++: Boosting cost aggregation with convolutions and transformers},
  author={Cho, Seokju and Hong, Sunghwan and Kim, Seungryong},
  journal={IEEE Transactions on Pattern Analysis and Machine Intelligence},
  volume={45},
  number={6},
  pages={7174--7194},
  year={2022},
  publisher={IEEE}
}

@inproceedings{hong2022cost,
  title={Cost aggregation with 4d convolutional swin transformer for few-shot segmentation},
  author={Hong, Sunghwan and Cho, Seokju and Nam, Jisu and Lin, Stephen and Kim, Seungryong},
  booktitle={European Conference on Computer Vision},
  pages={108--126},
  year={2022},
  organization={Springer}
}

@inproceedings{hong2021deep,
  title={Deep matching prior: Test-time optimization for dense correspondence},
  author={Hong, Sunghwan and Kim, Seungryong},
  booktitle={Proceedings of the IEEE/CVF international conference on computer vision},
  pages={9907--9917},
  year={2021}
}

@inproceedings{cho2024cat,
  title={Cat-seg: Cost aggregation for open-vocabulary semantic segmentation},
  author={Cho, Seokju and Shin, Heeseong and Hong, Sunghwan and Arnab, Anurag and Seo, Paul Hongsuck and Kim, Seungryong},
  booktitle={Proceedings of the IEEE/CVF Conference on Computer Vision and Pattern Recognition},
  pages={4113--4123},
  year={2024}
}

@inproceedings{an2025cross,
  title={Cross-view completion models are zero-shot correspondence estimators},
  author={An, Honggyu and Kim, Jin Hyeon and Park, Seonghoon and Jung, Jaewoo and Han, Jisang and Hong, Sunghwan and Kim, Seungryong},
  booktitle={Proceedings of the Computer Vision and Pattern Recognition Conference},
  pages={1103--1115},
  year={2025}
}

@article{kim2025seg4diff,
  title={Seg4Diff: Unveiling Open-Vocabulary Segmentation in Text-to-Image Diffusion Transformers},
  author={Kim, Chaehyun and Shin, Heeseong and Hong, Eunbeen and Yoon, Heeji and Arnab, Anurag and Seo, Paul Hongsuck and Hong, Sunghwan and Kim, Seungryong},
  journal={arXiv preprint arXiv:2509.18096},
  year={2025}
}

@article{han2025d,
  title={D\^{} 2USt3R: Enhancing 3D Reconstruction with 4D Pointmaps for Dynamic Scenes},
  author={Han, Jisang and An, Honggyu and Jung, Jaewoo and Narihira, Takuya and Seo, Junyoung and Fukuda, Kazumi and Kim, Chaehyun and Hong, Sunghwan and Mitsufuji, Yuki and Kim, Seungryong},
  journal={arXiv preprint arXiv:2504.06264},
  year={2025}
}

@article{jocher2021ultralytics,
  title={ultralytics/yolov5: v4. 0-nn. SiLU () activations, Weights \& Biases logging, PyTorch Hub integration},
  author={Jocher, Glenn and Stoken, Alex and Borovec, Jirka and Changyu, Liu and Hogan, Adam and Chaurasia, Ayush and Diaconu, Laurentiu and Ingham, Francisco and Colmagro, Adrien and Ye, Hu and others},
  journal={Zenodo},
  year={2021}
}

@inproceedings{leng2025repa,
  title={Repa-e: Unlocking vae for end-to-end tuning of latent diffusion transformers},
  author={Leng, Xingjian and Singh, Jaskirat and Hou, Yunzhong and Xing, Zhenchang and Xie, Saining and Zheng, Liang},
  booktitle={Proceedings of the IEEE/CVF International Conference on Computer Vision},
  pages={18262--18272},
  year={2025}
}

@inproceedings{li2025garf,
  title={Garf: Learning generalizable 3d reassembly for real-world fractures},
  author={Li, Sihang and Jiang, Zeyu and Chen, Grace and Xu, Chenyang and Tan, Siqi and Wang, Xue and Fang, Irving and Zyskowski, Kristof and McPherron, Shannon P and Iovita, Radu and others},
  booktitle={Proceedings of the IEEE/CVF International Conference on Computer Vision},
  pages={5711--5721},
  year={2025}
}

@article{sun2025rectified,
  title={Rectified point flow: Generic point cloud pose estimation},
  author={Sun, Tao and Zhu, Liyuan and Huang, Shengyu and Song, Shuran and Armeni, Iro},
  journal={arXiv preprint arXiv:2506.05282},
  year={2025}
}

@article{yu2024representation,
  title={Representation alignment for generation: Training diffusion transformers is easier than you think},
  author={Yu, Sihyun and Kwak, Sangkyung and Jang, Huiwon and Jeong, Jongheon and Huang, Jonathan and Shin, Jinwoo and Xie, Saining},
  journal={arXiv preprint arXiv:2410.06940},
  year={2024}
}

@article{sellan2022breaking,
  title={Breaking bad: A dataset for geometric fracture and reassembly},
  author={Sell{\'a}n, Silvia and Chen, Yun-Chun and Wu, Ziyi and Garg, Animesh and Jacobson, Alec},
  journal={Advances in Neural Information Processing Systems},
  volume={35},
  pages={38885--38898},
  year={2022}
}

@inproceedings{qi2025two,
  title={Two by two: Learning multi-task pairwise objects assembly for generalizable robot manipulation},
  author={Qi, Yu and Ju, Yuanchen and Wei, Tianming and Chu, Chi and Wong, Lawson LS and Xu, Huazhe},
  booktitle={Proceedings of the Computer Vision and Pattern Recognition Conference},
  pages={17383--17393},
  year={2025}
}

@article{zhou2023uni3d,
  title={Uni3d: Exploring unified 3d representation at scale},
  author={Zhou, Junsheng and Wang, Jinsheng and Ma, Baorui and Liu, Yu-Shen and Huang, Tiejun and Wang, Xinlong},
  journal={arXiv preprint arXiv:2310.06773},
  year={2023}
}

@article{hadgi2026patchalign3d,
  title={PatchAlign3D: Local Feature Alignment for Dense 3D Shape understanding},
  author={Hadgi, Souhail and Gong, Bingchen and Sundararaman, Ramana and Pierson, Emery and Li, Lei and Wonka, Peter and Ovsjanikov, Maks},
  journal={arXiv preprint arXiv:2601.02457},
  year={2026}
}

@inproceedings{ma2025find,
  title={Find any part in 3d},
  author={Ma, Ziqi and Yue, Yisong and Gkioxari, Georgia},
  booktitle={Proceedings of the IEEE/CVF International Conference on Computer Vision},
  pages={7818--7827},
  year={2025}
}

@article{zhang2025concerto,
  title={Concerto: Joint 2d-3d self-supervised learning emerges spatial representations},
  author={Zhang, Yujia and Wu, Xiaoyang and Lao, Yixing and Wang, Chengyao and Tian, Zhuotao and Wang, Naiyan and Zhao, Hengshuang},
  journal={arXiv preprint arXiv:2510.23607},
  year={2025}
}

@inproceedings{wu2025sonata,
  title={Sonata: Self-supervised learning of reliable point representations},
  author={Wu, Xiaoyang and DeTone, Daniel and Frost, Duncan and Shen, Tianwei and Xie, Chris and Yang, Nan and Engel, Jakob and Newcombe, Richard and Zhao, Hengshuang and Straub, Julian},
  booktitle={Proceedings of the Computer Vision and Pattern Recognition Conference},
  pages={22193--22204},
  year={2025}
}

@article{liu2023openshape,
  title={Openshape: Scaling up 3d shape representation towards open-world understanding},
  author={Liu, Minghua and Shi, Ruoxi and Kuang, Kaiming and Zhu, Yinhao and Li, Xuanlin and Han, Shizhong and Cai, Hong and Porikli, Fatih and Su, Hao},
  journal={Advances in neural information processing systems},
  volume={36},
  pages={44860--44879},
  year={2023}
}

@article{huang2020dgl,
  title={Generative 3d part assembly via dynamic graph learning},
  author={Zhan, Guanqi and Fan, Qingnan and Mo, Kaichun and Shao, Lin and Chen, Baoquan and Guibas, Leonidas J and Dong, Hao and others},
  journal={Advances in Neural Information Processing Systems},
  volume={33},
  pages={6315--6326},
  year={2020}
}

@article{jones2020shapeassembly,
  title={Shapeassembly: Learning to generate programs for 3d shape structure synthesis},
  author={Jones, R Kenny and Barton, Theresa and Xu, Xianghao and Wang, Kai and Jiang, Ellen and Guerrero, Paul and Mitra, Niloy J and Ritchie, Daniel},
  journal={ACM Transactions on Graphics (TOG)},
  volume={39},
  number={6},
  pages={1--20},
  year={2020},
  publisher={ACM New York, NY, USA}
}

@inproceedings{besl1992icp,
  title={Method for registration of 3-D shapes},
  author={Besl, Paul J and McKay, Neil D},
  booktitle={Sensor fusion IV: control paradigms and data structures},
  volume={1611},
  pages={586--606},
  year={1992},
  organization={Spie}
}

@inproceedings{lamb2023fantastic,
  title={Fantastic Breaks: A Dataset of Paired 3D Scans of Real-World Broken Objects and Their Complete Counterparts},
  author={Lamb, Nikolas and Palmer, Cameron and Molloy, Benjamin and Banerjee, Sean and Banerjee, Natasha Kholgade},
  booktitle=CVPR,
  year={2023}
}

@inproceedings{wu2023leveraging,
  title={Leveraging SE (3) Equivariance for Learning 3D Geometric Shape Assembly},
  author={Wu, Ruihai and Tie, Chenrui and Du, Yushi and Zhao, Yan and Dong, Hao},
  booktitle=ICCV,
  year={2023}
}

@article{lu2023jigsaw,
  title={Jigsaw: Learning to assemble multiple fractured objects},
  author={Lu, Jiaxin and Sun, Yifan and Huang, Qixing},
  journal={Advances in Neural Information Processing Systems},
  volume={36},
  pages={14969--14986},
  year={2023}
}

@misc{falcon2019lightning,
  author       = {William Falcon and The PyTorch Lightning team},
  title        = {PyTorch Lightning},
  year         = {2019},
  url          = {https://github.com/PyTorchLightning/pytorch-lightning}
}

@inproceedings{lee2024pmtr,
  author    = {Lee, Nahyuk and Min, Juhong and Lee, Junha and Kim, Seungwook and Lee, Kanghee and Park, Jaesik and Cho, Minsu},
  title     = {3D Geometric Shape Assembly via Efficient Point Cloud Matching},
  booktitle = {Proceedings of the International Conference on Machine Learning (ICML)},
  year      = {2024},
}

@inproceedings{loshchilovdecoupled,
  title={Decoupled Weight Decay Regularization},
  author={Loshchilov, Ilya and Hutter, Frank},
  booktitle=ICLR,
  year={2019}
}

@article{harada2016proposal,
  title={Proposal of a shape adaptive gripper for robotic assembly tasks},
  author={Harada, Kensuke and Nagata, Kazuyuki and Rojas, Juan and Ramirez-Alpizar, Ixchel G and Wan, Weiwei and Onda, Hiromu and Tsuji, Tokuo},
  journal={Advanced Robotics},
  volume={30},
  number={17-18},
  pages={1186--1198},
  year={2016},
  publisher={Taylor \& Francis}
}

@inproceedings{zakka2020form2fit,
  title={Form2fit: Learning shape priors for generalizable assembly from disassembly},
  author={Zakka, Kevin and Zeng, Andy and Lee, Johnny and Song, Shuran},
  booktitle={2020 IEEE International Conference on Robotics and Automation (ICRA)},
  pages={9404--9410},
  year={2020},
  organization={IEEE}
}

@inproceedings{mcbride2003archaeological,
  title={Archaeological fragment reconstruction using curve-matching},
  author={McBride, Jonah C and Kimia, Benjamin B},
  booktitle=CVPRW,
  year={2003},
}

@inproceedings{son2013axially,
  title={Axially symmetric 3D pots configuration system using axis of symmetry and break curve},
  author={Son, Kilho and Almeida, Eduardo B and Cooper, David B},
  booktitle=CVPR,
  year={2013}
}

@article{qi2017pointnet++,
  title={Pointnet++: Deep hierarchical feature learning on point sets in a metric space},
  author={Qi, Charles Ruizhongtai and Yi, Li and Su, Hao and Guibas, Leonidas J},
  journal={Advances in neural information processing systems},
  volume={30},
  year={2017}
}

@inproceedings{wang2024puzzlefusion++,
  author    = {Wang, Zhengqing and Chen, Jiacheng and Furukawa, Yasutaka},
  title     = {PuzzleFusion++: Auto-agglomerative 3D Fracture Assembly by Denoise and Verify},
  booktitle   = ICLR,
  year      = {2025}
}

@article{cortes2012cka,
  title={Algorithms for learning kernels based on centered alignment},
  author={Cortes, Corinna and Mohri, Mehryar and Rostamizadeh, Afshin},
  journal={The Journal of Machine Learning Research},
  volume={13},
  pages={795--828},
  year={2012},
  publisher={JMLR. org}
}

@article{singh2025irepa,
  title={What matters for Representation Alignment: Global Information or Spatial Structure?},
  author={Singh, Jaskirat and Leng, Xingjian and Wu, Zongze and Zheng, Liang and Zhang, Richard and Shechtman, Eli and Xie, Saining},
  journal={arXiv preprint arXiv:2512.10794},
  year={2025}
}

@inproceedings{peebles2023dit,
  title={Scalable diffusion models with transformers},
  author={Peebles, William and Xie, Saining},
  booktitle={Proceedings of the IEEE/CVF international conference on computer vision},
  pages={4195--4205},
  year={2023}
}

@article{wu2025geometry,
  title={Geometry forcing: Marrying video diffusion and 3d representation for consistent world modeling},
  author={Wu, Haoyu and Wu, Diankun and He, Tianyu and Guo, Junliang and Ye, Yang and Duan, Yueqi and Bian, Jiang},
  journal={arXiv preprint arXiv:2507.07982},
  year={2025}
}

@article{wang2025repa,
  title={REPA Works Until It Doesn't: Early-Stopped, Holistic Alignment Supercharges Diffusion Training},
  author={Wang, Ziqiao and Zhao, Wangbo and Zhou, Yuhao and Li, Zekai and Liang, Zhiyuan and Shi, Mingjia and Zhao, Xuanlei and Zhou, Pengfei and Zhang, Kaipeng and Wang, Zhangyang and others},
  journal={arXiv preprint arXiv:2505.16792},
  year={2025}
}

@article{wu2025representation,
  title={Representation entanglement for generation: Training diffusion transformers is much easier than you think},
  author={Wu, Ge and Zhang, Shen and Shi, Ruijing and Gao, Shanghua and Chen, Zhenyuan and Wang, Lei and Chen, Zhaowei and Gao, Hongcheng and Tang, Yao and Yang, Jian and others},
  journal={arXiv preprint arXiv:2507.01467},
  year={2025}
}

@article{yoo2025structure,
  title={Structure-from-Sherds++: Robust Incremental 3D Reassembly of Axially Symmetric Pots from Unordered and Mixed Fragment Collections},
  author={Yoo, Seong Jong and Liu, Sisung and Arshad, Muhammad Zeeshan and Kim, Jinhyeok and Kim, Young Min and Aloimonos, Yiannis and Fermuller, Cornelia and Joo, Kyungdon and Kim, Jinwook and Hong, Je Hyeong},
  journal={arXiv preprint arXiv:2502.13986},
  year={2025}
}

@article{chaudhuri2011probabilistic,
  title={Probabilistic reasoning for assembly-based 3D modeling},
  author={Chaudhuri, Siddhartha and Kalogerakis, Evangelos and Guibas, Leonidas and Koltun, Vladlen},
  journal={ACM Transactions on Graphics (TOG)},
  volume={30},
  number={4},
  pages={1--10},
  year={2011},
  publisher={ACM New York, NY, USA}
}

@inproceedings{li2024category,
  title={Category-level multi-part multi-joint 3d shape assembly},
  author={Li, Yichen and Mo, Kaichun and Duan, Yueqi and Wang, He and Zhang, Jiequan and Shao, Lin},
  booktitle={Proceedings of the IEEE/CVF Conference on Computer Vision and Pattern Recognition},
  pages={3281--3291},
  year={2024}
}

@inproceedings{lu2025survey,
  title={A survey on computational solutions for reconstructing complete objects by reassembling their fractured parts},
  author={Lu, Jiaxin and Liang, Yongqing and Han, Huijun and Hua, Jiacheng and Jiang, Junfeng and Li, Xin and Huang, Qixing},
  booktitle={Computer Graphics Forum},
  volume={44},
  pages={e70081},
  year={2025},
  organization={Wiley Online Library}
}

@article{rousseeuw1987silhouettes,
  title={Silhouettes: a graphical aid to the interpretation and validation of cluster analysis},
  author={Rousseeuw, Peter J},
  journal={Journal of computational and applied mathematics},
  volume={20},
  pages={53--65},
  year={1987},
  publisher={Elsevier}
}

@inproceedings{huang1997image,
  title={Image indexing using color correlograms},
  author={Huang, Jing and Kumar, S Ravi and Mitra, Mandar and Zhu, Wei-Jing and Zabih, Ramin},
  booktitle={Proceedings of IEEE computer society conference on Computer Vision and Pattern Recognition},
  pages={762--768},
  year={1997},
  organization={IEEE}
}

@article{oquab2023dinov2,
  title={Dinov2: Learning robust visual features without supervision},
  author={Oquab, Maxime and Darcet, Timoth{\'e}e and Moutakanni, Th{\'e}o and Vo, Huy and Szafraniec, Marc and Khalidov, Vasil and Fernandez, Pierre and Haziza, Daniel and Massa, Francisco and El-Nouby, Alaaeldin and others},
  journal={arXiv preprint arXiv:2304.07193},
  year={2023}
}

@inproceedings{radford2021learning,
  title={Learning transferable visual models from natural language supervision},
  author={Radford, Alec and Kim, Jong Wook and Hallacy, Chris and Ramesh, Aditya and Goh, Gabriel and Agarwal, Sandhini and Sastry, Girish and Askell, Amanda and Mishkin, Pamela and Clark, Jack and others},
  booktitle={International conference on machine learning},
  pages={8748--8763},
  year={2021},
  organization={PmLR}
}

@inproceedings{yu2022pointbert,
  title={Point-bert: Pre-training 3d point cloud transformers with masked point modeling},
  author={Yu, Xumin and Tang, Lulu and Rao, Yongming and Huang, Tiejun and Zhou, Jie and Lu, Jiwen},
  booktitle={Proceedings of the IEEE/CVF conference on computer vision and pattern recognition},
  pages={19313--19322},
  year={2022}
}

@inproceedings{wu2024ptv3,
  title={Point transformer v3: Simpler faster stronger},
  author={Wu, Xiaoyang and Jiang, Li and Wang, Peng-Shuai and Liu, Zhijian and Liu, Xihui and Qiao, Yu and Ouyang, Wanli and He, Tong and Zhao, Hengshuang},
  booktitle={Proceedings of the IEEE/CVF conference on computer vision and pattern recognition},
  pages={4840--4851},
  year={2024}
}

@inproceedings{zhai2023siglip,
  title={Sigmoid loss for language image pre-training},
  author={Zhai, Xiaohua and Mustafa, Basil and Kolesnikov, Alexander and Beyer, Lucas},
  booktitle={Proceedings of the IEEE/CVF international conference on computer vision},
  pages={11975--11986},
  year={2023}
}

@article{dosovitskiy2020vit,
  title={An image is worth 16x16 words: Transformers for image recognition at scale},
  author={Dosovitskiy, Alexey and Beyer, Lucas and Kolesnikov, Alexander and Weissenborn, Dirk and Zhai, Xiaohua and Unterthiner, Thomas and Dehghani, Mostafa and Minderer, Matthias and Heigold, Georg and Gelly, Sylvain and others},
  journal={arXiv preprint arXiv:2010.11929},
  year={2020}
}

@article{xiang2017posecnn,
  title={Posecnn: A convolutional neural network for 6d object pose estimation in cluttered scenes},
  author={Xiang, Yu and Schmidt, Tanner and Narayanan, Venkatraman and Fox, Dieter},
  journal={arXiv preprint arXiv:1711.00199},
  year={2017}
}

@inproceedings{hodan2018bop,
  title={Bop: Benchmark for 6d object pose estimation},
  author={Hodan, Tomas and Michel, Frank and Brachmann, Eric and Kehl, Wadim and GlentBuch, Anders and Kraft, Dirk and Drost, Bertram and Vidal, Joel and Ihrke, Stephan and Zabulis, Xenophon and others},
  booktitle={Proceedings of the European conference on computer vision (ECCV)},
  pages={19--34},
  year={2018}
}

@inproceedings{xu2025spaformer,
  title={Spaformer: Sequential 3d part assembly with transformers},
  author={Xu, Boshen and Zheng, Sipeng and Jin, Qin},
  booktitle={2025 International Conference on 3D Vision (3DV)},
  pages={1317--1327},
  year={2025},
  organization={IEEE}
}

@inproceedings{deitke2023objaverse,
  title={Objaverse: A universe of annotated 3d objects},
  author={Deitke, Matt and Schwenk, Dustin and Salvador, Jordi and Weihs, Luca and Michel, Oscar and VanderBilt, Eli and Schmidt, Ludwig and Ehsani, Kiana and Kembhavi, Aniruddha and Farhadi, Ali},
  booktitle={Proceedings of the IEEE/CVF conference on computer vision and pattern recognition},
  pages={13142--13153},
  year={2023}
}

@article{chang2015shapenet,
  title={Shapenet: An information-rich 3d model repository},
  author={Chang, Angel X and Funkhouser, Thomas and Guibas, Leonidas and Hanrahan, Pat and Huang, Qixing and Li, Zimo and Savarese, Silvio and Savva, Manolis and Song, Shuran and Su, Hao and others},
  journal={arXiv preprint arXiv:1512.03012},
  year={2015}
}

@article{fu20213d,
  title={3d-future: 3d furniture shape with texture},
  author={Fu, Huan and Jia, Rongfei and Gao, Lin and Gong, Mingming and Zhao, Binqiang and Maybank, Steve and Tao, Dacheng},
  journal={International Journal of Computer Vision},
  volume={129},
  number={12},
  pages={3313--3337},
  year={2021},
  publisher={Springer}
}

@inproceedings{collins2022abo,
  title={Abo: Dataset and benchmarks for real-world 3d object understanding},
  author={Collins, Jasmine and Goel, Shubham and Deng, Kenan and Luthra, Achleshwar and Xu, Leon and Gundogdu, Erhan and Zhang, Xi and Vicente, Tomas F Yago and Dideriksen, Thomas and Arora, Himanshu and others},
  booktitle={Proceedings of the IEEE/CVF conference on computer vision and pattern recognition},
  pages={21126--21136},
  year={2022}
}
\end{document}